\documentclass[11pt]{article}

\usepackage[utf8]{inputenc} 
\usepackage[T1]{fontenc}    
\usepackage{hyperref}

\usepackage[table]{xcolor}
\usepackage{url}

\usepackage{tikz}
\usepackage{tikz-qtree}
\usetikzlibrary{shapes.geometric,arrows}
\usepackage{tango-colors}
\usepackage{pythonhighlight}

\tikzstyle{edge}=[->, >=stealth', shorten <=2pt, shorten >=2pt, semithick]
\tikzstyle{node}=[inner sep=0.1cm, semithick, circle]
\tikzstyle{p1}=[fill=orange1,draw=orange3]
\tikzstyle{p2}=[fill=skyblue1,draw=skyblue3]
\tikzstyle{p1color}=[orange3]
\tikzstyle{p2color}=[skyblue3]

\usepackage{pgfplots}
\usepgfplotslibrary{fillbetween}
\pgfplotsset{compat=1.13}

\usepackage{algorithm}
\usepackage{algpseudocode}

\usepackage{booktabs}
\usepackage{amsmath}
\usepackage{bm}
\usepackage{amssymb}
\usepackage{amsthm}

\newtheorem{thm}{Theorem}

\usepackage{csquotes}
\usepackage{listings}
\newcommand{\R}{\mathbb{R}}

\newcommand{\enc}{f}
\newcommand{\dec}{g}
\newcommand{\Enc}{\phi}
\newcommand{\Dec}{\psi}

\newcommand{\tre}[1]{\hat{#1}}

\usepackage[a4paper, left=3cm, right=2.5cm, top=3cm, bottom=3cm]{geometry}

\usepackage{authblk}
\usepackage{natbib}
\usepackage{doi}

\newcommand{\arxivswitch}[2]{#2}

\begin{document}

\title{ast2vec: Utilizing Recursive Neural Encodings of Python Programs}

\author[1]{Benjamin Paaßen}
\author[1]{Jessica McBroom}
\author[2]{Bryn Jeffries}
\author[1]{Irena Koprinska}
\author[1]{Kalina Yacef}
\affil[1]{School of Computer Science\\The University of Sydney}
\affil[2]{Grok Learning}

\date{This work is a preprint, provided by the authors, and has been submitted to the Journal of Educational Datamining} 
\pagestyle{myheadings}
\markright{Preprint version under consideration at the Journal of Educational Datamining}

\maketitle

\maketitle

\begin{abstract}
Educational datamining involves the application of datamining techniques to student activity. However, in the context of computer programming, many datamining techniques can not be applied because they expect vector-shaped input whereas computer programs have the form of syntax trees. In this paper, we present ast2vec, a neural network that maps Python syntax trees to vectors and back, thereby facilitating datamining on computer programs as well as the interpretation of datamining results. Ast2vec has been trained on almost half a million programs of novice programmers and is designed to be applied across learning tasks \emph{without re-training}, meaning that users can apply it without any need for (additional) deep learning. We demonstrate the generality of ast2vec in three settings: First, we provide example analyses using ast2vec on a classroom-sized dataset, involving visualization, student motion analysis, clustering, and outlier detection, including two novel analyses, namely a
progress-variance-projection and a dynamical systems analysis. Second, we consider the ability of ast2vec to recover the original syntax tree from its vector representation on the training data and two further large-scale programming datasets. Finally, we evaluate the predictive capability of a simple linear regression on top of ast2vec, obtaining similar results to techniques that work directly on syntax trees.
We hope ast2vec can augment the educational datamining toolbelt
by making analyses of computer programs easier, richer, and more efficient.

{\parindent0pt
\textbf{Keywords:} computer science education, computer programs, word embeddings, representation learning, neural networks, visualization, program vectors
}
\end{abstract}

\section{Introduction}

Techniques for analyzing and utilizing student programs have been the focus of much recent research in 
computer science education. Such techniques have included hint systems to provide automated feedback to 
students \citep{Paassen2018JEDM,Piech2015,Price2019,Rivers2017}, as well as visualization and search 
approaches to help teachers understand student behavior \citep{McBroom2018,Nguyen2014}. Considering that 
programming is a key skill in many fields, including science, technology, engineering and mathematics 
\citep{Denning2017,McCracken2001,Wiles2009}, and considering the difficulty students have with learning 
programming \citep{Denning2017,Lahtinen2005,Qian2017,Robins2003}, developing and expanding the range of 
available techniques to improve educational outcomes is of great importance.

Unfortunately, computer programs are particularly difficult to analyze for two main reasons. First, 
programs come in the form of raw code or syntax trees (after compilation), which few datamining 
techniques are equipped to  handle. Instead, one first has to represent programs differently to turn them
into valid input for data  mining techniques \citep{Paassen2018JEDM}. Second, the space of possible
programs grows combinatorially with program length and most programs are never written by a student,
even fewer more than once \citep{Paassen2018JEDM,Rivers2012,Rivers2017}. Accordingly, one needs to
abstract from meaningless differences between programs to shrink the space and, thus, make it easier to
handle with less risk of overfitting.

Several prior works have addressed both the representation and the abstraction step, often in conjunction.
For example, \citet{Rivers2012} have suggested semantically motivated transformations of syntax trees 
to remove syntactic variations that have no semantic influence (such as unreachable code or 
direction of binary operators). \citet{Peddycord2014} suggest to represent programs by their output 
rather than their syntax. Similarly, \citet{Paassen2016} suggest to represent programs by their execution 
behavior. \citet{Gulwani2018} as well as \citet{Gross2014} perform a clustering of programs to achieve a 
representation in terms of a few discrete clusters. We point to the 'related work' section and to the 
review of \citet{McBroom2019} for a more comprehensive list. We also note that many of the possible 
abstraction and representation steps are not opposed but can be combined to achieve better 
results \citep{McBroom2019}.

The arguably most popular representation of computer programs are pairwise tree edit distances \citep{Zhang1989}. For example, \citet{Mokbel2013}, 
\citet{Paassen2018JEDM}, \citet{Price2017}, and \citet{Rivers2017} have applied variations of the standard tree edit distance for processing programs. Edit distances have the advantage 
that they do not only quantify distance between programs, they also specify which parts of the code have 
to be changed to transform one program into another, which can be utilized to construct hints 
\citep{Paassen2018JEDM,Price2017,Rivers2017}. Additionally, many datamining techniques for visualization, 
clustering, regression, and classification can deal with input in terms of pairwise distances 
\citep{Pekalska2005,Hammer2010}.
Still, a vast majority of techniques can not. For example, of 126 methods contained in the Python library scikit-learn\footnote{Taken from this list 
\url{https://scikit-learn.org/stable/modules/classes.html}.}, only 24 can natively deal with pairwise 
distances, eight further methods can deal with kernels, which require additional 
transformations \citep{Gisbrecht2015}, and 94 only work with vector-shaped input. As such, having a 
vector-formed representation of computer programs enables a much wider range of analysis techniques. 
Further, a distance-based representation depends on a database of reference programs to compare it to. The computational complexity required for 
analyzing a program thus scales at least linearly with the size of the database. By contrast, a 
parametric model with vector-shaped input can perform operations independent of the size of the 
training data. Finally, a representation in terms of pairwise distances 
tries to solve the representation problem for computer programs every time anew for each new learning 
task. Conceptually, it appears more efficient to share representational knowledge across tasks. In this 
paper, we aim to achieve just that: To find a mapping from computer programs to vectors in Euclidean space 
(and back) that generalizes across learning tasks and thus solves the representation problem ahead of 
time so that we, as educational dataminers, only need to add a simple model specific to the analysis task 
we wish to solve. In other words, we wish to achieve for computer programs what word embeddings like 
word2vec \citep{Mikolov2013} or GloVe \citep{Pennington2014} offer for natural language processing: a 
re-usable component that solves the representation problem of programs such that subsequent research can 
concentrate on other datamining challenges. Our technique is by no means meant to supplant all existing
representations. It merely is supposed to be another tool on the toolbelt for educational datamining
in computer science education.

We acknowledge that we are not the first to attempt such a vector-shaped representation. Early work has
utilized hand-engineered features such as program length or cyclomatic complexity \citep{Truong2004}.
More recent approaches check whether certain tree patterns occur within a given syntax tree, which gives
rise to a vector with one dimension per tree pattern that is zero if the syntax tree does not contain
the pattern and nonzero if it does \citep{Nguyen2014,Zhi2018,Zimmerman2015}.
One can also derive a vectorial representation from pairwise edit distances using linear algebra 
(although the dependence on a database remains) \citep{Paassen2018JEDM}. Finally, prior works have trained
neural networks to map programs to vectors \citep{Piech2015ICML,Alon2019}. However, all these
representations lose a desirable property of other representations named above, namely the ability to
return from a program's representation to the original program.
For example, while it is simple to compute the length of a program, it is impossible
to identify the original program just from its length. However, such an inverse translation is
crucial for the interpretability of a result. For example, if I want to provide a hint to a student,
it would not help the student much if I tell them that they should increase the length of their program.
Instead, one would rather like to know which parts of their code they have to change.

So how does one achieve a mapping from computer programs to vectors that can be inverted and still 
generalizes across tasks? The solution appears to lie in recent works on autoencoders for tree-structured
data \citep{Chen2018,Dai2018,Kusner2017,Paassen2021}, which train an encoder $\Enc$ from trees to vectors and a 
decoder $\Dec$ from vectors back to trees such that $\Enc(\Dec(x))$ is similar to $x$ for as many trees 
$x$ as possible from the training data. While training such a model requires large amounts of data and a
lot of computation time (as is usual in deep learning), our hope is that a trained model can then be
used as a representation and abstraction device without any further need for deep learning. Instead, one
can just apply vector-based datamining techniques, such as any of the methods in scikit-learn, to the 
vector representations of computer programs and even transform any vector back into a program if desired.

In this paper, we present ast2vec, an
autoencoder for trees trained on 
almost half a million Python programs written by novice programmers that encodes beginner Python programs 
as $256$-dimensional vectors in Euclidean space \emph{and} decodes vectors as syntax trees. Both encoding 
and decoding are guided by the grammar of the Python programming language, thereby ensuring that decoded 
trees are syntactically correct. The model and its source code is completely open-source and is available 
at \url{https://gitlab.com/bpaassen/ast2vec/}.

The overarching question guiding our research is: Does ast2vec generalize across learning tasks 
and across different datamining techniques? To this end, we test ast2vec in three settings: First, we 
utilize ast2vec for a variety of analyses on a classroom-sized dataset, thereby showing 
the generality qualitatively. Second, we quantitatively investigate the autoencoding error of ast2vec 
on its training data and on two similarly large datasets of novice programs from subsequent years. In all cases, we observe that small trees usually get autoencoded correctly whereas errors may occur for larger trees.
Finally, we investigate the performance of a simple linear regression model on top of ast2vec for 
predicting the next step of students in a learning task and observe that we achieve comparable results to 
established techniques while being both more space- and more time-efficient.

This paper is set out as follows: Section~\ref{sec:tutorial} begins by utilizing ast2vec for a variety of
datamining analyses on a classroom-sized dataset to demonstrate its broad applicability qualitatively.
Section~\ref{sec:method} then specifies the details of ast2vec, 
including how it was trained and constructed, as well as the details of two novel analysis techniques used in Section~\ref{sec:tutorial}, namely progress-variance projections and dynamic analyses. Next, 
Section~\ref{sec:evaluation} provides an evaluation of ast2vec on large-scale datasets, including an 
investigation of its autoencoding performance and predictive capabilities. Finally, Section~\ref{sec:related_work} describes related work in more detail and Section~\ref{sec:conclusion} concludes 
with a summary of the main ideas of the paper and a discussion of future directions.

\section{Example analyses using ast2vec}
\label{sec:tutorial}

Before introducing the details of ast2vec, we perform example analysis to investigate how well the vector representation of ast2vec generalizes across methods. In particular, we perform four example analyses using ast2vec on a classroom-sized 
dataset: 1) visualizing student work and progress on a task,
2) modeling student behavior,
3) clustering student programs, and
4) outlier detection.

Note that we have used a synthethic dataset for this example analysis because we wanted to make all of the code and data used in this section publicly available (it can be found here: \url{https://gitlab.com/bpaassen/ast2vec/}).
However, we have not limited our analyses to synthetic data and, indeed, in Section \ref{sec:evaluation} we will discuss experiments on large-scale, real datasets.

\subsection{Dataset Construction}

To construct the dataset used in these example analyses, we manually simulated ten students attempting to solve the following
introductory programming task:

\begin{quote}
Write a program that asks the user, \enquote{What are your favourite animals?}. If their response is \enquote{Quokkas}, the program should print \enquote{Quokkas are the best!}. If their response is anything else, the program should print \enquote{I agree, x are great animals.}, where x is their response.
\end{quote}

We simulated the development process by starting with an empty program, writing a partial
program, receiving feedback from automatic test cases for this partial program, and revising
the program in response to the test feedback until a correct solution was reached. The reference solution for the task is the following:
\begin{center}
\begin{minipage}{0.7\linewidth}
\begin{python}
	1		x = input("What are your favourite animals? ")
	2		if x == "Quokkas":
	3		    print("Quokkas are the best!")
	4		else:
	5		    print(f"I agree, {x} are great animals.")
\end{python}
\end{minipage}
\end{center}

Our four simulated unit tests checked 1) whether the first line of output was \enquote{What are your favourite animals?}, and 2-4) whether the remaining output was correct for the inputs \enquote{Quokkas}, \enquote{Koalas} and \enquote{Echidnas}, respectively.

Overall, the dataset contains $58$ (partial) programs, and these form $N = 25$ unique syntax trees
after compilation. Since ast2vec converts program trees to $n=256$ dimensional vectors, this means 
the data matrix $\bm{X} \in \R^{N \times n}$
after encoding all programs is of size $N \times n = 25 \times 256$.

\subsection{Application 1: Visualizing Student Work and Progress}
As a first example application of ast2vec, we consider a visualization of student progress from the empty program to the correct solution. The purpose of such a
visualization is to give educators a concise overview of typical paths towards the goal, where
students tend to struggle, as well as how many typical strategies exist \citep{McBroom2021}.

In order to create such a visualization, we combine ast2vec with a second technique that we call \textit{progress-variance projection}. Our key idea is to construct a linear projection from 256 dimensions to two dimensions, where the first axis captures the progress from the empty program towards the goal and the second axis captures as much of the remaining variance between programs as possible. A detailed description of this approach is given in Section~\ref{sec:analysis_1_construction}.

\begin{figure}
\begin{center}
\begin{tikzpicture}
\begin{axis}[view={0}{90}, xlabel={progress}, ylabel={variance},
xmin = 0, xmax = 1, ymin = -0.1, ymax = 1,
enlarge x limits=0.05, enlarge y limits=0.05,
width=9cm, height=6.5cm]

\node[circle, inner sep = 0.158114cm, fill=aluminium4, draw=aluminium6] (x0) at (axis cs:0, 0) {};
\node[circle, inner sep = 0.15cm, fill=aluminium4, draw=aluminium6] (x1) at (axis cs:0.469296, 0.575221) {};
\node[circle, inner sep = 0.1cm, fill=aluminium4, draw=aluminium6] (x2) at (axis cs:0.641942, 0.492747) {};
\node[circle, inner sep = 0.05cm, fill=aluminium4, draw=aluminium6] (x3) at (axis cs:0.807819, 0.158199) {};
\node[circle, inner sep = 0.132288cm, fill=aluminium4, draw=aluminium6] (x4) at (axis cs:1, -3.43881e-16) {};
\node[circle, inner sep = 0.0707107cm, fill=aluminium4, draw=aluminium6] (x5) at (axis cs:0.688732, 0.445487) {};
\node[circle, inner sep = 0.05cm, fill=aluminium4, draw=aluminium6] (x6) at (axis cs:0.876653, 0.407659) {};
\node[circle, inner sep = 0.111803cm, fill=aluminium4, draw=aluminium6] (x7) at (axis cs:0.611764, 0.847813) {};
\node[circle, inner sep = 0.05cm, fill=aluminium4, draw=aluminium6] (x8) at (axis cs:0.665497, 0.99769) {};
\node[circle, inner sep = 0.05cm, fill=aluminium4, draw=aluminium6] (x9) at (axis cs:0.714276, 0.765656) {};
\node[circle, inner sep = 0.0707107cm, fill=aluminium4, draw=aluminium6] (x10) at (axis cs:0.778177, 0.428875) {};
\node[circle, inner sep = 0.111803cm, fill=aluminium4, draw=aluminium6] (x11) at (axis cs:0.821145, 0.245713) {};
\node[circle, inner sep = 0.0707107cm, fill=aluminium4, draw=aluminium6] (x12) at (axis cs:0.444297, 0.138948) {};
\node[circle, inner sep = 0.0866025cm, fill=aluminium4, draw=aluminium6] (x13) at (axis cs:0.997112, -0.122418) {};
\node[circle, inner sep = 0.0707107cm, fill=aluminium4, draw=aluminium6] (x14) at (axis cs:0.881619, 0.201604) {};
\node[circle, inner sep = 0.05cm, fill=aluminium4, draw=aluminium6] (x15) at (axis cs:0.99918, 0.107273) {};
\node[circle, inner sep = 0.05cm, fill=aluminium4, draw=aluminium6] (x16) at (axis cs:0.912179, 0.27201) {};
\node[circle, inner sep = 0.05cm, fill=aluminium4, draw=aluminium6] (x17) at (axis cs:0.73415, 0.288822) {};
\node[circle, inner sep = 0.0707107cm, fill=aluminium4, draw=aluminium6] (x18) at (axis cs:0.509801, 0.927955) {};
\node[circle, inner sep = 0.05cm, fill=aluminium4, draw=aluminium6] (x19) at (axis cs:0.58455, 0.491723) {};
\node[circle, inner sep = 0.0707107cm, fill=aluminium4, draw=aluminium6] (x20) at (axis cs:0.901792, 0.337555) {};
\node[circle, inner sep = 0.05cm, fill=aluminium4, draw=aluminium6] (x21) at (axis cs:1.01953, -0.0930224) {};
\node[circle, inner sep = 0.05cm, fill=aluminium4, draw=aluminium6] (x22) at (axis cs:0.8925, 0.471583) {};
\node[circle, inner sep = 0.05cm, fill=aluminium4, draw=aluminium6] (x23) at (axis cs:0.904299, 0.424788) {};
\node[circle, inner sep = 0.05cm, fill=aluminium4, draw=aluminium6] (x24) at (axis cs:0.955138, 0.466791) {};

\path[edge, fill = orange1, draw = orange3]
(x0) edge (x1)
(x1) edge (x2)
(x2) edge (x3)
(x3) edge (x4)
;

\path[edge, fill = skyblue1, draw = skyblue3]
(x0) edge[bend left=10] (x1)
(x1) edge[loop above, looseness=8] (x1)
(x1) edge (x5)
(x5) edge (x6)
(x6) edge (x4)
;

\path[edge, fill = scarletred1, draw = scarletred3]
(x0) edge[bend left=20] (x1)
(x1) edge (x7)
(x7) edge[loop above, looseness=8] (x7)
(x7) edge (x8)
(x8) edge (x9)
(x9) edge (x10)
(x10) edge (x11)
(x11) edge[loop above, looseness=8] (x11)
(x11) edge[loop above, looseness=10] (x11)
(x11) edge (x4)
;

\path[edge, fill = plum1, draw = plum3]
(x0) edge (x12)
(x12) edge[loop above, looseness=8] (x12)
(x12) edge (x2)
(x2) edge[loop above, looseness=8] (x2)
(x2) edge (x5)
(x5) edge (x13)
(x13) edge[loop above, looseness=8] (x13)
;

\path[edge, fill = butter1, draw = butter3]
(x0) edge[bend left=30] (x1)
(x1) edge[loop above, looseness=10] (x1)
(x1) edge[bend left=10] (x7)
(x7) edge (x14)
(x14) edge[loop above, looseness=8] (x14)
(x14) edge (x15)
(x15) edge (x4)
;

\path[edge, fill = chameleon1, draw = chameleon3]
(x0) edge[bend left=40] (x1)
(x1) edge[loop above, looseness=12] (x1)
(x1) edge[loop above, looseness=14] (x1)
(x1) edge[bend left=20] (x7)
(x7) edge (x2)
(x2) edge (x11)
(x11) edge (x16)
(x16) edge (x17)
;

\path[edge, fill = chocolate1, draw = chocolate3]
(x0) edge (x7)
(x7) edge (x18)
(x18) edge[loop above, looseness=8] (x18)
(x18) edge (x19)
(x19) edge (x20)
(x20) edge[loop above, looseness=8] (x20)
(x20) edge (x21)
(x21) edge (x13)
;

\path[edge, fill = orange1, draw = orange3]
(x0) edge (x10)
(x10) edge[bend left=10] (x11)
(x11) edge[bend left=10] (x4)
;

\path[edge, fill = skyblue1, draw = skyblue3]
(x0) edge (x4)
;

\path[edge, fill = scarletred1, draw = scarletred3]
(x0) edge (x22)
(x22) edge (x23)
(x23) edge (x24)
(x24) edge (x4)
;

\end{axis}

\begin{scope}[shift={(8,-0.05)}]
\node[inner sep=0.1cm, fill=white,draw=black,align=left, below right] (x1_prog) at (0,5) {
\begin{lstlisting}
print('<string>')
\end{lstlisting}
};
\node[inner sep=0.1cm, fill=white,draw=black,align=left, below right] (x7_prog) at (0,4) {
\begin{lstlisting}
input('<string>')
print('<string>')
\end{lstlisting}
};
\node[inner sep=0.1cm, fill=white,draw=black,align=left, below right] (x4_prog) at (0,2.5) {
\begin{lstlisting}
x = input('<string>')
if x == '<string>':
  print('<string>')
else:
  print('<string>')
\end{lstlisting}
};
\end{scope}

\path[edge]
(x1_prog.west) edge[out=180,in=90] (x1)
(x4_prog.west) edge[out=180,in=0] (x4)
(x7_prog.west) edge[out=180,in=0] (x7);

\end{tikzpicture}
\end{center}
\caption{A progress-variance plot of the example task. The special points $(0,0)$ and $(1,0)$
correspond to the empty program and the reference solution, respectively. The size of points
corresponds to their frequency in the data. Arrows indicate student motion. Different students
are plotted in different colors.}
\label{fig:traces}
\end{figure}
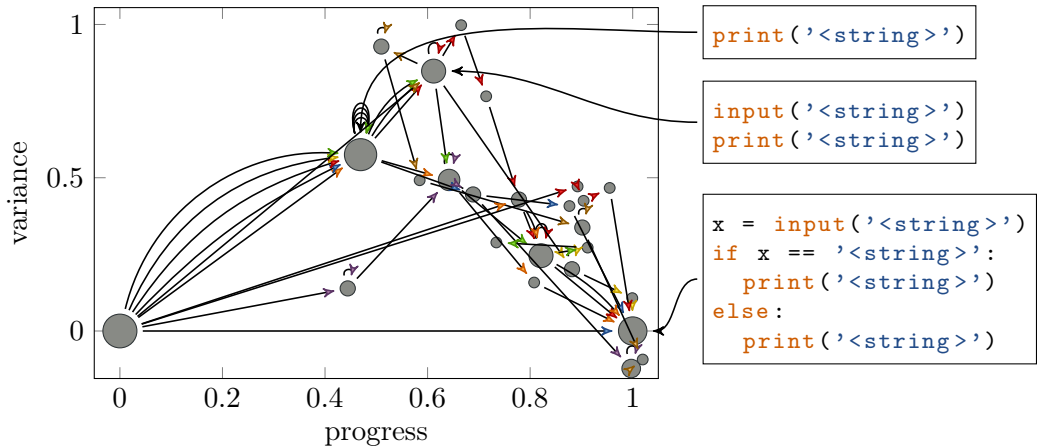

Figure~\ref{fig:traces} shows the result of applying this process to the sample data. 
Each circle in the plot represents a unique syntax tree in the data, with larger circles representing more frequent trees and the three most common trees shown on the right. Note that $(0,0)$ represents the empty program and $(1,0)$ represents the solution. In addition, the arrows indicate student movement between programs, with different colors representing different students. More specifically, an arrow is drawn from vector $\vec x$ to $\vec y$ if the student made a program submission corresponding to vector $\vec x$ followed by a submission corresponding to vector $\vec y$.

Without the axis labels, this plot is similar to the interaction networks suggested
by \citet{Barnes2016}, which have already been shown to provide useful insights into student learning \citep{McBroom2018}. However, in contrast to interaction networks, our progress-variance projection additionally maps vectors to meaningful locations in space. In particular, x-axis corresponds to the (linear) progress toward the solution, whereas the y-axis corresponds to variation between student programs that is orthogonal to the progress direction. This can provide an intuitive overview of the types of programs students submit and how they progress through the exercise.

Additionally, the plot reveals to us that most syntax trees only occur a single time and that
student programs differ a lot on a fine-grained level, especially close to the solution around $(0.8, 0.4)$.
However, while exact repetitions of syntax trees are rare, the coarse-grained motion 
of students seems to be consistent, following an arc from the origin
to the correct solution. We will analyze this motion in more detail in the next section.

Finally, we notice that multiple students tend to get stuck
at a program which only does a single function call (corresponding to point $(0.5, 0.55)$
in the plot). This stems from programs that have the correct syntax tree but not the right string
(e.g.\ due to typos). Accordingly, if students fix the string they still have the same syntax tree and, hence, the same location in the visualization.

In summary, this section demonstrates how ast2vec can be used to visualize student progress through a programming task, how we can interpret such a visualization, and how the vectorial output of ast2vec enabled us to develop a novel visualization technique in the first place.

\subsection{Application 2: Dynamical analysis}
\label{sec:dynamics}

In Figure~\ref{fig:traces}, we have observed that the high-level motion of students
appears to be consistent. In particular, all students seem to follow an arc-shaped
trajectory beginning at $(0,0)$, then rising in the $y$-axis, before approaching the
reference solution at $(1, 0)$. This raises the question: Can we capture this general
motion of students in a simple, dynamical system? The purpose of such a prediction could
be to provide next-step hints. If a student does not know how to continue, we can provide
a prediction how students would generally proceed, namely following the arc in the plot.
Then, we can use ast2vec to decode the predicted position back into a syntax tree and use
an edit distance to construct the hint, similar to prior techniques \citep{Paassen2018JEDM,Price2017,Rivers2017}.

In Section~\ref{sec:analysis_2_construction}, we discuss
one technique to perform such a prediction. In particular, we learn a linear
dynamical system that has the correct solution to the
task as a unique stable attractor.

\begin{figure}
\begin{center}
\begin{tikzpicture}
\begin{axis}[view={0}{90}, xlabel={progress}, ylabel={variance},
enlarge x limits=0.05, enlarge y limits=0.05]
\addplot3+[no marks, p1, -stealth', semithick, quiver={u=\thisrow{u}, v=\thisrow{v}, scale arrows=0.3}]
table[x=x, y=y, col sep=tab] {pca_grid_dynsys.csv};

\node[node, p2] (x0) at (axis cs:0,0) {};
\node[node, p2] (x1) at (axis cs:0.30,0.24) {};
\node[node, p2] (x2) at (axis cs:0.49,0.36) {};
\node[node, p2] (x3) at (axis cs:0.62,0.40) {};
\node[node, p2] (x4) at (axis cs:0.72,0.35) {};
\node[node, p2] (x5) at (axis cs:0.80,0.27) {};
\node[node, p2] (x6) at (axis cs:0.86,0.20) {};
\node[node, p2] (x7) at (axis cs:0.91,0.14) {};

\path[edge, p2]
(x0) edge (x1)
(x1) edge (x2)
(x2) edge (x3)
(x3) edge (x4)
(x4) edge (x5)
(x5) edge (x6)
(x6) edge (x7);

\end{axis}

\begin{scope}[shift={(7.5,-0.05)}]
\node[inner sep=0.1cm, fill=white,draw=black,align=left, below right] (x1_prog) at (0,5.75) {
\begin{lstlisting}
x = input('<string>')
\end{lstlisting}
};

\node[inner sep=0.1cm, fill=white,draw=black,align=left, below right] (x2_prog) at (0,4.25) {
\begin{lstlisting}
x = input('<string>')
print('<string>')
\end{lstlisting}
};

\node[inner sep=0.1cm, fill=white,draw=black,align=left, below right] (x5_prog) at (0,2.5) {
\begin{lstlisting}
x = input('<string>')
if x == '<string>':
  print('<string>')
else:
  print('<string>')
\end{lstlisting}
};
\end{scope}

\path[edge]
(x1_prog.west) edge[out=180,in=90] (x1)
(x2_prog.west) edge[out=180,in=45] (x2)
(x5_prog.west) edge[out=180,in=-90] (x5);

\end{tikzpicture}
\end{center}
\caption{A linear dynamical system $f$, which has a single attractor at the
correct solution and approximates the motion of students through the space of programs
(orange). Arrows are scaled with factor 0.3 for better visibility. An example trace starting at the empty program and following the dynamical
system is shown in blue. Whenever the decoded program changes along the trace, we
show the code on the right. The coordinate system is the same as in Figure~\ref{fig:traces}.}
\label{fig:dynsys}
\end{figure}
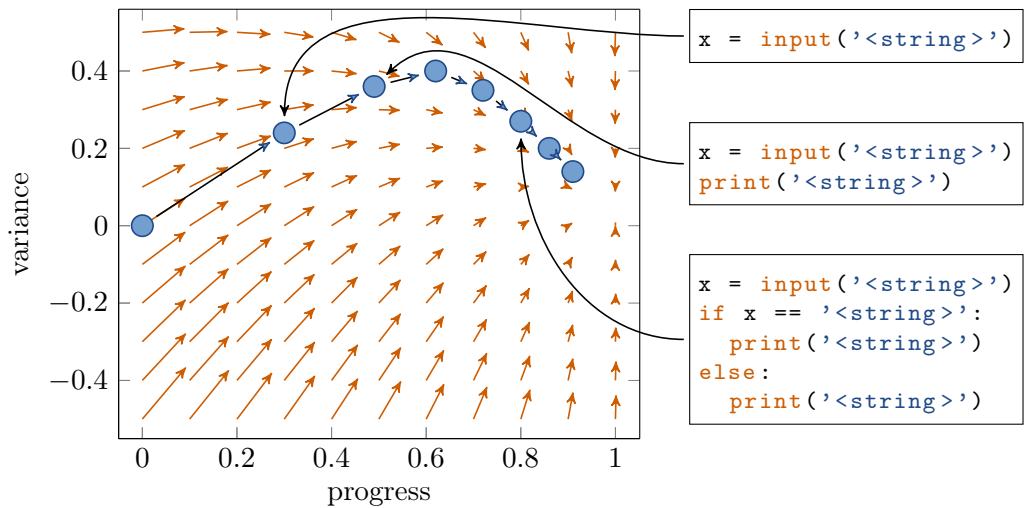

Figure~\ref{fig:dynsys} illustrates the dynamical system we obtain from our example data. In particular, the orange arrows
indicate how a point at the origin of the arrow would be moved by the dynamical system. As we can see, the dynamical system does indeed capture
the qualitative behavior we observe in Figure~\ref{fig:traces}, namely the arc-shaped
motion from the empty program towards the reference solution. We can verify this finding
by simulating a new trace that follows our dynamical system. In particular, we start at
the empty program $\vec x_0$ and then set $\vec x_{t+1} = f(x_t)$ until the location does
not change much anymore. The resulting motion is plotted
in blue. We further decode all steps $\vec x_t$ using ast2vec and inspect the resulting
syntax tree. These trees are shown in Figure~\ref{fig:dynsys} on the right.
We observe that the simulated trace corresponds to reasonable student motion, namely to
first ask the user for input, then add a single print statement, and finally to extend
the solution with an if-statement that can distinguish between the input 'Quokka'
and all other inputs.

\subsection{Application 3: Clustering}
\label{sec:clustering}

When inspecting Figure~\ref{fig:traces}, we also notice that student programs seem
to fall into clusters, namely one cluster around the correct solution,
one cluster to the top left of it, and further minor clusters. Finding clusters in programming
data is useful both for providing feedback based on clusters \citep{Gross2014} as well as
reducing data complexity by summarizing a big dataset of programs by a few representative
programs \citep{Glassman2015}.

Given that ast2vec represents our data as vectors, we can use any clustering technique.
In this case, we use a Gaussian mixture model, which also provides us with a notion of probability \citep{Dempster1977}.
In particular, a Gaussian mixture model expresses the data as a mixture of $K$ Gaussians
\begin{equation*}
p(\vec x) = \sum_{k=1}^K p(\vec x|k) \cdot p(k),
\end{equation*}
where $p(\vec x|k)$ is the standard Gaussian density. The mean, covariance, and prior $p(k)$
of each Gaussian are learned according to the data. After training, we assign data vectors
$\vec x_i$ to clusters by evaluating the posterior probability $p(k|\vec x_i)$ and assigning
the Gaussian $k$ with highest probability. Note that, prior to clustering, we perform a standard principal component analysis (PCA) because distances tend to degenerate for high dimensionality
and thus complicate distance-based clusterings \citep{Aggarwal2001}. We set the PCA to
preserve $95\%$ of the data variance, which yielded $12$ dimensions. Then, we cluster
the data via a Gaussian mixture model with $K = 4$ Gaussians.

\begin{figure}
\begin{center}
\begin{tikzpicture}
\begin{axis}[view={0}{90}, xlabel={progress}, ylabel={variance},
xmin = 0, xmax = 1, ymin = -0.1, ymax = 1,
enlarge x limits=0.05, enlarge y limits=0.05,
width=8cm, height=6cm]

\addplot[scatter/classes={%
0={mark=*,fill=orange1,draw=orange3},%
1={mark=*,fill=skyblue1,draw=skyblue3},%
2={mark=*,fill=butter1,draw=butter3},%
3={mark=*,fill=plum1,draw=plum3}},
scatter,only marks,mark size=2,
scatter src=explicit symbolic]
table[x=x,y=y,meta=c] {pca_clustering.csv};

\node[diamond, inner sep=0.1cm, semithick, fill=orange1, draw=orange3] (w0) at (axis cs:0.982517, 0.0452825) {};
\node[diamond, inner sep=0.1cm, semithick, fill=skyblue1, draw=skyblue3] (w1) at (axis cs:0.611977, 0.797146) {};
\node[diamond, inner sep=0.1cm, semithick, fill=butter1, draw=butter3] (w2) at (axis cs:0.59749, 0.61758) {};
\node[diamond, inner sep=0.1cm, semithick, fill=plum1, draw=plum3] (w3) at (axis cs:0.744286, 0.309667) {};

\node[circle, inner sep=0.08cm, semithick, draw=scarletred3] at (axis cs:0, 0) {};
\node[circle, inner sep=0.08cm, semithick, draw=scarletred3] at (axis cs:0.444297, 0.138948) {};
\node[circle, inner sep=0.08cm, semithick, draw=scarletred3] at (axis cs:0.73415, 0.288822) {};
\node[circle, inner sep=0.08cm, semithick, draw=scarletred3] at (axis cs:0.904299, 0.424788) {};
\node[circle, inner sep=0.08cm, semithick, draw=scarletred3] at (axis cs:0.955138, 0.466791) {};

\end{axis}

\node[inner sep=0.1cm, fill=white,draw=black,align=left, below right] (w0_prog) at (7,0.5) {
\begin{lstlisting}
x = input('<string>')
if x == '<string>':
  print(f('<string>' + x))
\end{lstlisting}
};
\node[inner sep=0.1cm, fill=white,draw=black,align=left, below right] (w3_prog) at (7,2.5) {
\begin{lstlisting}
x = input('<string>')
if x == '<string>':
  print('<string>')
\end{lstlisting}
};
\node[inner sep=0.1cm, fill=white,draw=black,align=left, below right] (w2_prog) at (7,4) {
\begin{lstlisting}
x = input('<string>')
print('<string>')
\end{lstlisting}
};
\node[inner sep=0.1cm, fill=white,draw=black,align=left, below right] (w1_prog) at (7,5.5) {
\begin{lstlisting}
input('<string>')
print('<string>')
\end{lstlisting}
};

\path[edge]
(w0_prog.west) edge[out=180,in=0] (w0)
(w1_prog.west) edge[out=180,in=0] (w1)
(w2_prog.west) edge[out=180,in=0] (w2)
(w3_prog.west) edge[out=180,in=0] (w3);

\end{tikzpicture}
\end{center}
\caption{A visualization of Gaussian mixture clusters of programs with color indicating
cluster assignment. Cluster centers are shown as diamonds. Outliers are highlighted with
a red ring. The coordinate system is the same as in Figure~\ref{fig:traces}.}
\label{fig:clusters}
\end{figure}
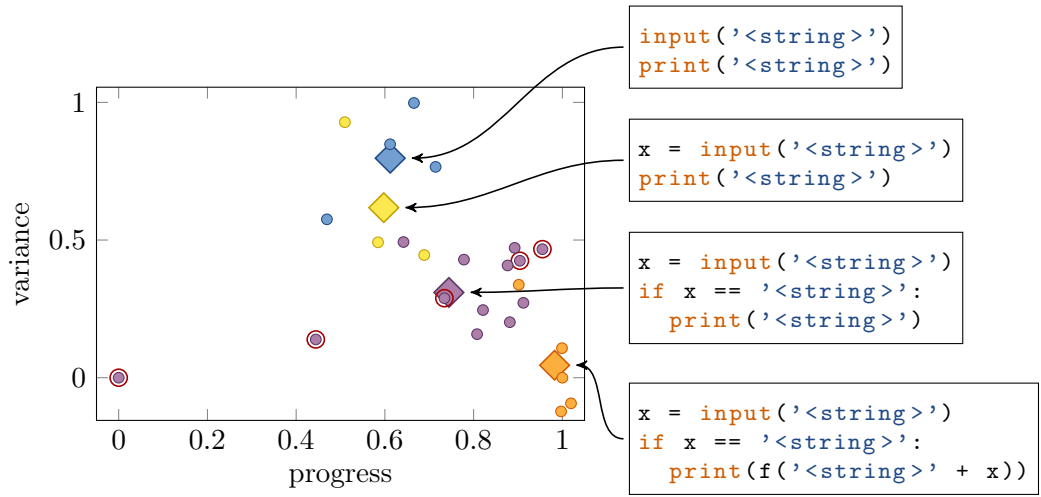

Figure~\ref{fig:clusters} shows the resulting clustering with clusters indicated by color
and cluster means highlighted as diamonds. We also decode all means back into syntax trees
using ast2vec. Indeed, we notice that the clustering roughly corresponds to the progress
through the task with the blue cluster mean containing an input and a print statement, the yellow
cluster mean storing the input in a variable, the purple cluster mean adding an if statement, and
the orange cluster mean printing out the user input.

\subsection{Application 4: Outlier Detection}
An additional benefit of a probability-based clustering is that we can use it directly to
detect \enquote{outliers}. Detecting such outliers can be useful because they lie outside the range
of usual data and thus may require special attention by educators.

Figure~\ref{fig:clusters} highlights all points with a
red circle that received particularly low probability $p(\vec x)$ by our Gaussian mixture model%
\footnote{We defined 'particularly low' as having a log probability of at most half the average
when normalizing by the least likely sample.}.
Unsurprisingly, one outlier is the empty program. However, the other outliers are more instructive. In particular, the outlier in the center of the plot corresponds to an input statement without
a print statement, which is an unusual path towards the solution. In this dataset,
it is more common to write the print statement first. The two outliers around $(0.9,0.45)$
correspond to the following program shape:
\begin{center}
\begin{minipage}{0.75\linewidth}
\begin{python}
1		print('What are your favourite animals? ')
2		animals = input()
3		if animals == 'Quokkas':
4		    print('Quokkas are the best!')
5		else:
6		    print(f'I agree, {animals} are great animals.')
\end{python}
\end{minipage}
\end{center}
Here, the question for the favourite animals is posed as a print statement and
the input is requested with a separate command, which is a misunderstanding of how
\pyth{input} works.

The final outlier is the program:
\begin{center}
\begin{minipage}{0.75\linewidth}
\begin{python}
1		animals = input('What are your favourite animals? ')
2		if animals == 'Koalas':
3		    print('I agree, Koalas are great animals.')
4		elif animals == 'Echidnas':
5 		    print('I agree, Echidnas are great animals.')
6		else:
7		    print('Quokkas are the best!')
\end{python}
\end{minipage}
\end{center}
This program does pass all our test cases but does not adhere to the \enquote{spirit} of the
task, because it does not generalize to new inputs. Such an outlier may instruct us that
we need to change our test cases to be more general, e.g.\ by using a hidden test with a case that is not known to the student.

This concludes our example data analyses using ast2vec. We note that further types of
data analysis are very well possible, e.g.\ to define an interaction network \citep{Barnes2016}
on top of a clustering.
In the remainder of this paper, we will explain how ast2vec works in more detail and evaluate
it on large-scale data.

\section{Methods}
\label{sec:method}

In this section, we describe the methods employed in this paper in more detail. In Section~\ref{sec:architecture}, we provide a summary of the autoencoder approach we used for ast2vec. In Section~\ref{sec:analysis_1_construction}, we describe the progress-variance projection that we used to generate 2D visualizations of student data. Finally, in Section~\ref{sec:analysis_2_construction}, we explain how to learn linear dynamical systems from student data.

\subsection{The Architecture of ast2vec}
\label{sec:architecture}

Our neural network ast2vec is an instance of a recursive tree grammar autoencoder model as proposed by \citet{Paassen2021}. In this section, we describe this approach as we used it for ast2vec. If readers are interested in the general approach (and more technical details), we recommend the original paper.

On a high level, ast2vec is a so-called
\emph{autoencoder}, i.e.\ a combination of an encoder $\Enc : \mathcal{X} \to \R^n$ from trees
to vectors and a decoder $\Dec : \R^n \to \mathcal{X}$ from vectors to trees, such that $\Dec(\Enc(\tre x))$
is equal to $\tre x$ for as many trees $\tre x$ as possible. 
Unfortunately, this problem has undesirable solutions if we only consider a finite training
dataset. In particular, let $\tre x_1, \ldots, \tre x_N \in \mathcal{X}$ be a training
dataset of trees. Then, we can set $\Enc(\tre x_i) = i$ and $\Dec(i) = \tre x_i$, i.e.\
we implement $\Enc$ and $\Dec$ as a simple lookup table that is only defined exactly
on the training data and does not generalize to any other tree.

\begin{figure}
\begin{center}
\begin{tikzpicture}
\begin{scope}[shift={(0,0.5)}]
\node[p1color] at (-0.25,0.2) {$\tre x$};
\node (tree1_out) at (0.3,0.4) {};

\node[node,p1] (x)  at (0,0) {};
\node[node,p1] (y1) at (-0.3,-0.5) {};
\node[node,p1] (y2) at (+0.3,-0.5) {};
\node[node,p1] (z)  at (+0.3,-1) {};

\path[edge, -,p1]
(x)  edge (y1) edge (y2)
(y2) edge (z);


\end{scope}

\begin{scope}[shift={(2,0.5)}]
\node[p2color] at (-0.25,0.2) {$\tre y$};
\node (tree2_in) at (0.4,-0.6) {};

\node[node,p2] (x)  at (0,0) {};
\node[node,p2] (y1) at (-0.3,-0.5) {};
\node[node,p2] (y2) at (+0.3,-0.5) {};
\node[node,p2] (z1) at (-0.6,-1) {};
\node[node,p2] (z2) at ( 0,-1) {};

\path[edge, -, p2]
(x)  edge (y1) edge (y2)
(y1) edge (z1) edge (z2);

\end{scope}

\begin{scope}[shift={(6,0)}]
\draw[semithick, black, ->, >=stealth'] (-1.5,0) -- (+1.5,0);
\draw[semithick, black, ->, >=stealth'] (0,-1.5) -- (0,+1.5);

\node[node,p1] (x1)  at (-1,0.8) {};
\node[p1color, above] at (x1.north) {$\vec x$};

\node[node,p2] (x1e) at (+0.4,0.6) {};
\node[p2color, above right] at (x1e.center) {$\vec x + \vec \epsilon$};

\path[edge, p1]
(tree1_out) edge[bend left] node[above, p1color] {$\Enc$} (x1);

\path[edge, fill=aluminium4, draw=aluminium6]
(x1) edge node[above, aluminium6] {$\vec \epsilon$} (x1e);

\path[edge, p2]
(x1e) edge[bend left] node[below, p2color] {$\Dec$} (tree2_in);

\end{scope}
\end{tikzpicture}
\end{center}
\caption{A high-level illustration of the (denoising) autoencoder framework.
Syntax tree $\tre x$ (left) gets encoded as a vector $\vec x = \Enc(\tre x)$ (right).
We then add noise $\vec \epsilon$ and decode back to a tree
$\tre y = \Dec(\vec x + \vec \epsilon)$. If $\tre x \neq \tre y$, we adjust the parameters
of both encoder and decoder to increase the probability that $\vec x + \vec \epsilon$
is decoded into $\tre x$ instead of $\tre y$.}
\label{fig:vae}
\end{figure}
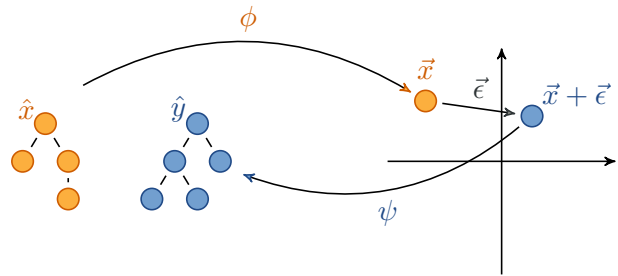

A remedy against this issue is the variational autoencoding (VAE) framework of
\citet{Kingma2013}.
In this framework, we add a small amount of Gaussian noise to the encoding vector before decoding it
again to a program (refer also to Figure~\ref{fig:vae} for an illustration).
Because this noise is different every time, the encoder-decoder pair must be
able to generalize at least slightly outside the training data. Additionally, VAEs ensure that the
distribution of vectors remains close to a standard normal distribution, which in turn means that
the encodings cannot degenerate to large values.
In more detail, VAEs rephrase the entire autoencoding problem in a probabilistic fashion, where
we replace the deterministic decoder with a probability distribution $p_\Dec(\tre x|\vec x)$
of vector $\vec x$ getting decoded to tree $\tre x$. We then wish to maximize the probability
that a syntax tree gets autoencoded correctly even after adding a noise vector $\vec \epsilon$
to the encoding vector; and we wish to keep the noisy code vectors standard normally distributed.
The precise optimization problem becomes
\begin{equation}
\min_{\substack{\Enc : \mathcal{X} \to \R^n \\ \Dec : \R^n \to \mathcal{X}}} \quad
\sum_{i=1}^N -\log\Big[p_\Dec\big( \tre x_i \big| \Enc(\tre x_i) + \epsilon_i \big)\Big]
+ \beta \cdot D_{KL}\Big(\Enc(\tre x_i) + \epsilon_i\Big), \label{eq:vae}
\end{equation}
where $\epsilon_i$ is a Gaussian noise vector that is generated randomly in every
round of training, $\beta$ is a hyper-parameter to regulate how smooth we want our coding
space to be, and $D_{KL}\Big(\Enc(\tre x_i) + \epsilon_i\Big)$ is the Kullback-Leibler divergence between 
the distribution of the noisy neural code $\Enc(\tre x_i) + \epsilon_i$ and the standard
normal distribution - i.e.\ it punishes if the code distribution becomes too different from a standard 
normal distribution.

In the next paragraphs, we describe the encoder $\Enc$, the decoder $\Dec$, the probability
distribution $p_\Dec$, and the training scheme for ast2vec.

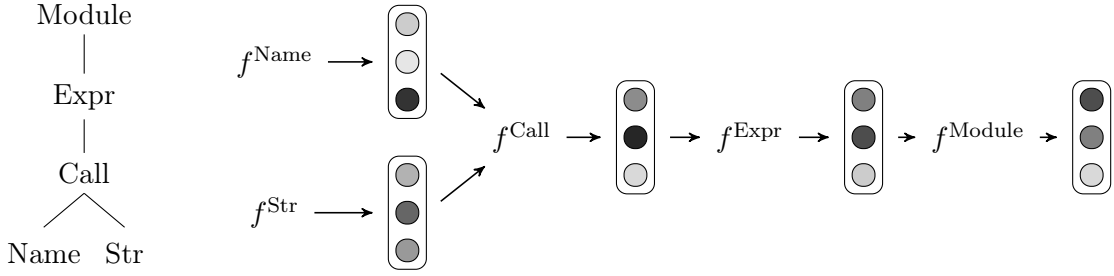
\begin{figure}
\begin{center}
\begin{tikzpicture}

\begin{scope}[shift={(3,+1.5)}]

\Tree [.Module [.Expr [.Call
{$\strut$Name}
{$\strut$Str}
] ] ]
\end{scope}

\node (fname) at (5.5,1) {$\enc^\text{Name}$};
\node (fstr)  at (5.5,-1) {$\enc^\text{Str}$};

\begin{scope}[shift={(7,1)}]
\node (phiname_in)  at (0,0) {};
\node (phiname_out) at (0.5,0) {};

\draw[rounded corners] (0,-0.75) rectangle (0.5,+0.75);
\draw[rounded corners, draw, fill=black!80] (0.1,-0.65) rectangle (0.4,-0.35);
\draw[rounded corners, draw, fill=black!10] (0.1,-0.15) rectangle (0.4,0.15);
\draw[rounded corners, draw, fill=black!20] (0.1,+0.35) rectangle (0.4,0.65);
\end{scope}

\begin{scope}[shift={(7,-1)}]
\node (phistr_in)  at (0,0) {};
\node (phistr_out) at (0.5,0) {};

\draw[rounded corners] (0,-0.75) rectangle (0.5,+0.75);
\draw[rounded corners, draw, fill=black!40] (0.1,-0.65) rectangle (0.4,-0.35);
\draw[rounded corners, draw, fill=black!60] (0.1,-0.15) rectangle (0.4,0.15);
\draw[rounded corners, draw, fill=black!30] (0.1,+0.35) rectangle (0.4,0.65);
\end{scope}

\node (fcall) at (8.75,0) {$\enc^\text{Call}$};

\begin{scope}[shift={(10,0)}]
\node (phicall_in)  at (0,0) {};
\node (phicall_out) at (0.5,0) {};

\draw[rounded corners] (0,-0.75) rectangle (0.5,+0.75);
\draw[rounded corners, draw, fill=black!15] (0.1,-0.65) rectangle (0.4,-0.35);
\draw[rounded corners, draw, fill=black!85] (0.1,-0.15) rectangle (0.4,0.15);
\draw[rounded corners, draw, fill=black!45] (0.1,+0.35) rectangle (0.4,0.65);
\end{scope}

\node (fexpr) at (11.75,0) {$\enc^\text{Expr}$};

\begin{scope}[shift={(13,0)}]
\node (phiexpr_in)  at (0,0) {};
\node (phiexpr_out) at (0.5,0) {};

\draw[rounded corners] (0,-0.75) rectangle (0.5,+0.75);
\draw[rounded corners, draw, fill=black!20] (0.1,-0.65) rectangle (0.4,-0.35);
\draw[rounded corners, draw, fill=black!70] (0.1,-0.15) rectangle (0.4,0.15);
\draw[rounded corners, draw, fill=black!50] (0.1,+0.35) rectangle (0.4,0.65);
\end{scope}

\node (fmod) at (14.75,0) {$\enc^\text{Module}$};

\begin{scope}[shift={(16,0)}]
\node (phimod_in)  at (0,0) {};

\draw[rounded corners] (0,-0.75) rectangle (0.5,+0.75);
\draw[rounded corners, draw, fill=black!15] (0.1,-0.65) rectangle (0.4,-0.35);
\draw[rounded corners, draw, fill=black!50] (0.1,-0.15) rectangle (0.4,0.15);
\draw[rounded corners, draw, fill=black!70] (0.1,+0.35) rectangle (0.4,0.65);
\end{scope}

\path[edge]
(fname) edge (phiname_in)
(fstr)  edge (phistr_in)
(phiname_out) edge (fcall)
(phistr_out)  edge (fcall)
(fcall) edge (phicall_in)
(phicall_out) edge (fexpr)
(fexpr) edge (phiexpr_in)
(phiexpr_out) edge (fmod)
(fmod) edge (phimod_in);

\end{tikzpicture}
\end{center}
\caption{An illustration of the encoding process into a vector with $n = 3$ dimensions.
We first use the Python compiler to transform the program
\lstinline!print('Hello, world!')!
into the abstract syntax tree Module(Expr(Call(Name, Str))) (left).
Then, from left to right, we apply the recursive encoding scheme,
where arrows represent the direction of computation and traffic light-elements
represent three-dimensional vectors, where color indicates the value of the vector
in that dimension.}
\label{fig:encoding}
\end{figure}

\paragraph{Encoder:} The first step of our encoding is to use the Python compiler to generate an
abstract syntax tree for the program. Now, let $x(\tre y_1, \ldots, \tre y_K)$ be such an abstract
syntax tree, where $x$ is the root syntactic element and $\tre y_1, \ldots,$ $\tre y_K$ are its $K$
child subtrees. For an example of such a syntax tree, refer to Figure~\ref{fig:encoding} (left).
Our encoder $\Enc$ is then defined as follows.
\begin{equation}
\Enc\Big(x(\tre y_1, \ldots, \tre y_K)\Big) := \enc^x\Big(\Enc(\tre y_1), \ldots, \Enc(\tre y_K) \Big),
\label{eq:enc}
\end{equation}
where $\enc^x : \R^{K \times n} \to \R^n$ is a function that takes the vectors
$\Enc(\tre y_1), \ldots, \Enc(\tre y_K)$ for all children as input and returns a vector
for the entire tree, including the syntactic element $x$ and all its children. Because
Equation~\ref{eq:enc} is recursively defined, we also call our encoding \emph{recursive}.
Figure~\ref{fig:encoding} shows an example encoding for the program \lstinline{print('Hello, world!')}
with $n = 3$ dimensions. We first use the Python compiler to translate this program into
the abstract syntax tree Module(Expr(Call(Name, Str))) and then apply our recursive
encoding scheme. In particular, recursively applying Equation~\ref{eq:enc} to this tree yields the
expression $\enc^\text{Module}(\enc^\text{Expr}(\enc^\text{Call}(\enc^\text{Name}(), \enc^\text{Str}())))$.
We can evaluate this expression by first computing the leaf terms $\enc^\text{Name}()$ and
$\enc^\text{Str}()$, which do not require any inputs because they have no children.
This yields two vectors, one representing Name and one representing Str, respectively.
Next, we feed these two vectors into the encoding function $\enc^\text{Call}$, which
transforms them into a vector code of the subtree Call(Name, Str). We feed this vector
code into $\enc^\text{Expr}$, whose output we feed into $\enc^\text{Module}$, which in turn
yields the overall vector encoding for the tree. Note that our computation follows the structure
of the tree bottom-up, where each encoding function receives exactly as many inputs as
it has children.

A challenge in this scheme is that we have to know the number of children $K$ of each
syntactic element $x$ in advance to construct a function $\enc^x$.
Fortunately, the grammar of the programming language%
\footnote{The entire grammar for the Python language can be found at this link:
\url{https://docs.python.org/3/library/ast.html}.}
tells us how many children are permitted
for each syntactic element. For example, an \lstinline{if} element in the
Python language has three children, one for the condition, one for the code that gets executed
if the condition evaluates to \lstinline{True} (the `then' branch), and one for the code that gets
executed if the condition evaluates to \lstinline{False} (the `else' branch).

Leaves, like Str or Name, are an important special case. Since these have no children, their
encoding is a constant vector $\enc^x \in \R^n$. This also means that we encode all possible strings
as the same vector (the same holds for all different variable names or all different numbers).
Incorporating content information of variables in the encoding is a topic for future research.

Another important special case are lists. For example, a Python program is defined as a list
of statements of arbitrary length. In our scheme, we encode a list by adding up all vectors inside
it. The empty list is represented as a zero vector.

Note that, up to this point, our procedure is entirely general and has nothing to do with neural
nets. Our approach becomes a (recursive) neural network because we implement 
the encoding functions $\enc^x$ as neural networks.
In particular, we use a simple single-layer feedforward network for the encoding function $\enc^x$ of the syntactic element $x$:
\begin{equation}
\enc^x(\vec y_1, \ldots, \vec y_K) = \tanh\Big(\bm{U}^x_1 \cdot \vec y_1 + \ldots + \bm{U}^x_K \cdot \vec y_K + \vec b^x\Big), \label{eq:net}
\end{equation}
where $\bm{U}^x_k \in \R^{n \times n}$ is the matrix that decides what information flows from
the $k$th child to its parent and $\vec b^x \in \R^n$ represents the syntactic element $x$.
These $\bm{U}$ matrices and the $\vec b$ vectors are the parameters of the encoder that need to
be learned during training.

Importantly, this architecture is still relatively simple. If one would aim to optimize autoencoding performance, one could imagine implementing $\enc^x$ instead with more advanced neural networks, such as a Tree-LSTM \citep{Chen2018,Dai2018,Tai2015}. This is a topic for future research.

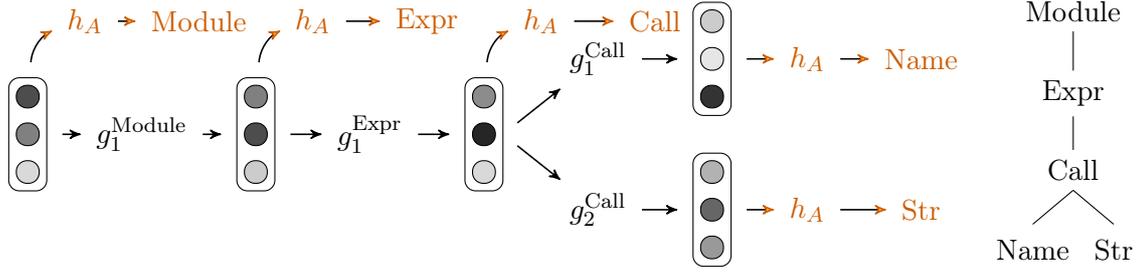
\begin{figure}
\begin{center}
\begin{tikzpicture}

\begin{scope}[shift={(0,0)}]
\node (phimod_out) at (0.5,0) {};
\node (phimod_north) at (0.25,0.75) {};

\draw[rounded corners] (0,-0.75) rectangle (0.5,+0.75);
\draw[rounded corners, draw, fill=black!15] (0.1,-0.65) rectangle (0.4,-0.35);
\draw[rounded corners, draw, fill=black!50] (0.1,-0.15) rectangle (0.4,0.15);
\draw[rounded corners, draw, fill=black!70] (0.1,+0.35) rectangle (0.4,0.65);

\node[orange3] (hMod) at (1,1.5) {$h_A$};
\node[orange3] (mod)  at (2.5,1.5) {Module};

\node (gMod) at (1.75,0) {$\dec_1^\text{Module}$};
\end{scope}

\begin{scope}[shift={(3,0)}]
\node (phiexpr_in)  at (0,0) {};
\node (phiexpr_out) at (0.5,0) {};
\node (phiexpr_north) at (0.25,0.75) {};

\draw[rounded corners] (0,-0.75) rectangle (0.5,+0.75);
\draw[rounded corners, draw, fill=black!20] (0.1,-0.65) rectangle (0.4,-0.35);
\draw[rounded corners, draw, fill=black!70] (0.1,-0.15) rectangle (0.4,0.15);
\draw[rounded corners, draw, fill=black!50] (0.1,+0.35) rectangle (0.4,0.65);

\node[orange3] (hExpr) at (1,1.5) {$h_A$};
\node[orange3] (expr)  at (2.5,1.5) {Expr};

\node (gExpr) at (1.75,0) {$\dec_1^\text{Expr}$};
\end{scope}

\begin{scope}[shift={(6,0)}]
\node (phicall_in)  at (0,0) {};
\node (phicall_out) at (0.5,0) {};
\node (phicall_north) at (0.25,0.75) {};

\draw[rounded corners] (0,-0.75) rectangle (0.5,+0.75);
\draw[rounded corners, draw, fill=black!15] (0.1,-0.65) rectangle (0.4,-0.35);
\draw[rounded corners, draw, fill=black!85] (0.1,-0.15) rectangle (0.4,0.15);
\draw[rounded corners, draw, fill=black!45] (0.1,+0.35) rectangle (0.4,0.65);

\node[orange3] (hCall) at (1,1.5) {$h_A$};
\node[orange3] (call)  at (2.5,1.5) {Call};

\node (gCall1) at (1.75,+1) {$\dec_1^\text{Call}$};
\node (gCall2) at (1.75,-1) {$\dec_2^\text{Call}$};
\end{scope}

\begin{scope}[shift={(9,1)}]
\node (phiname_in)  at (0,0) {};
\node (phiname_out) at (0.5,0) {};
\node (phiname_north) at (0.25,0.75) {};

\draw[rounded corners] (0,-0.75) rectangle (0.5,+0.75);
\draw[rounded corners, draw, fill=black!80] (0.1,-0.65) rectangle (0.4,-0.35);
\draw[rounded corners, draw, fill=black!10] (0.1,-0.15) rectangle (0.4,0.15);
\draw[rounded corners, draw, fill=black!20] (0.1,+0.35) rectangle (0.4,0.65);

\node[orange3] (hName) at (1.5,0) {$h_A$};
\node[orange3] (name)  at (3,0) {Name};
\end{scope}

\begin{scope}[shift={(9,-1)}]
\node (phistr_in)  at (0,0) {};
\node (phistr_out) at (0.5,0) {};
\node (phistr_north) at (0.25,0.75) {};

\draw[rounded corners] (0,-0.75) rectangle (0.5,+0.75);
\draw[rounded corners, draw, fill=black!40] (0.1,-0.65) rectangle (0.4,-0.35);
\draw[rounded corners, draw, fill=black!60] (0.1,-0.15) rectangle (0.4,0.15);
\draw[rounded corners, draw, fill=black!30] (0.1,+0.35) rectangle (0.4,0.65);

\node[orange3] (hStr) at (1.5,0) {$h_A$};
\node[orange3] (str)  at (3,0) {Str};
\end{scope}

\path[edge, fill=orange1, draw=orange3]
(phimod_north) edge[bend left] (hMod)
(hMod) edge (mod)
(phiexpr_north) edge[bend left] (hExpr)
(hExpr) edge (expr)
(phicall_north) edge[bend left] (hCall)
(hCall) edge (call)
(phiname_out) edge (hName)
(hName) edge (name)
(phistr_out) edge (hStr)
(hStr) edge (str);

\path[edge]
(phimod_out) edge (gMod)
(gMod) edge (phiexpr_in)
(phiexpr_out) edge (gExpr)
(gExpr) edge (phicall_in)
(phicall_out) edge (gCall1) edge (gCall2)
(gCall1) edge (phiname_in)
(gCall2) edge (phistr_in);

\begin{scope}[shift={(14,+1.5)}]

\Tree [.Module [.Expr [.Call
{$\strut$Name}
{$\strut$Str}
] ] ]
\end{scope}

\end{tikzpicture}
\end{center}
\caption{An illustration of the decoding process of a vector with $n = 3$ dimensions into
an abstract syntax tree. We first plug the current vector into a scoring function $h$, based
on which we select the current syntactic element $x$ (shown in orange). After this step, we feed
the current vector into decoding functions $\dec^x_1, \ldots, \dec^x_K$, one for each child a
syntactic element $x$ expects. This yields the next vectors for which the same process is
applied again until all remaining vectors decode into leaf elements.}
\label{fig:decoding}
\end{figure}

\paragraph{Decoding:} To decode a vector $\vec x$ recursively back into a syntax tree
$x(\tre y_1, \ldots, \tre y_K)$, we have to make $K+1$ decisions. First, we have to decide
the syntactic element $x$. Then, we have to decide the vector codes $\vec y_1, \ldots, \vec y_K$
for each child. For the first decision we set up a function $h_A$ which computes a numeric score
$h_A(x|\vec x)$ for each possible syntactic element $x$ from the vector $\vec x$.
We then select the syntactic element $x$ with the highest score. The $A$ in the index of $h$
refers to the fact that we guide our syntactic decision by the Python grammar. In particular,
we only allow syntactic elements to be chosen that are allowed by the current grammar symbol $A$.
All non-permitted elements receive a score of $-\infty$. For simplicity, we do not discuss the
details of grammar rules here but point the interested reader to \citet{Paassen2021}.

Once we know the syntactic element, the Python grammar tells us the number of children $K$, i.e.\
how many child codes we need to generate. Accordingly, we use $K$ decoding functions $\dec^x_k : \R^n \to \R^n$
which tell us the vector code for each child based on the parent code $\vec x$.
The precise definition of the decoding procedure is as follows.
\begin{align}
\Dec(\vec x) = &x\Big( \Dec(\vec y_1), \ldots, \Dec(\vec y_k) \Big) & \text{where} \label{eq:dec}\\
x = &\arg\max_y h_A(y|\vec x) \qquad \text{and} \quad \vec y_k = \dec^x_k(\vec x) &\text{ for all $k \in \{1, \ldots, K\}$} \notag
\end{align}
In other words, we first use $h_A$ to choose the current syntactic element $x$ with maximum
score $h_A(x|\vec x)$, use the decoding functions $g^x_1, \ldots, g^x_K$ to compute the
neural codes for all children, and proceed recursively to decode child subtrees.

An example of the decoding process is shown in Figure~\ref{fig:decoding}. As input, we receive
some vector $\vec x$, which we first feed into the scoring function $h_A$. As the Python grammar
requires that each program begins with a Module, the only score higher than $-\infty$ is 
$h(\text{Module}|\vec x)$, meaning that the root of our generated tree is Module.
Next, we feed the vector $\vec x$ into the decoding function $\dec^\text{Module}_1$, yielding
a vector $\dec^\text{Module}_1(\vec x)$ to be decoded into the child subtree. With this vector,
we re-start the process, i.e.\ we feed the vector $\dec^\text{Module}_1(\vec x)$ into our
scoring function $h_A$, which this time selects the syntactic element Expr. We then generate the
vector representing the child of Expr as $\vec y = \dec^\text{Expr}_1(\dec^\text{Module}_1(\vec x))$.
For this vector, $h_A$ selects Call, which expects two children. We obtain the vectors for these
two children as $\dec^\text{Call}_1(\vec y)$ and $\dec^\text{Call}_2(\vec y)$. For these vectors,
$h_A$ selects Name and Str, respectively. Neither of these has children, such that the process
stops, leaving us with the tree Module(Expr(Call(Name, Str))).

Again, we note the special case of lists. To decode a list, we use a similar scheme, where we
let $h_A$ make a binary choice whether to continue the list or not, and use the decoding function
$\dec^x_k$ to decide the code for the next list element.

We implement the decoding functions $g^x_k$ with single-layer
feedforward neural networks as in Equation~\ref{eq:net}. A special case are the decoding
functions for list-shaped children, which we implement as recurrent neural networks, namely gated recurrent units \citep{Cho2014}. Again, we note that one could choose to implement all decoding functions $g^x_k$ as recurrent networks, akin to the decoding scheme suggested by \citep{Chen2018}. Here, we opt for a simple version and leave the extension to future work.

For the scoring function $h_A$, we use a linear layer with $n$ inputs and as
many outputs as there are syntactic elements in the programming language.
Importantly, this choice process can also be modelled in a probabilistic fashion, where the
syntactic element $x$ is chosen with probability
\begin{equation}
p(x|\vec x) = \frac{\exp\big[h_A(x|\vec x)\big]}{\sum_y \exp\big[h_A(y|\vec x)\big]},
\end{equation}
i.e.\ $x$ is sampled according to a softmax distribution with weights $h_A(x|\vec x)$.
The probability $p_\Dec(\tre x|\vec x)$ from the optimization problem~\ref{eq:vae} is
then defined as the product over all these probabilities during decoding. In other words,
$p_\Dec(\tre x|\vec x)$ is the probability that the tree $\tre x$ is sampled via
decoder $\Dec$ when receiving $\vec x$ as input.

We note in passing that ast2vec only decodes to syntax trees and does not include variable
names, numbers, or strings. However, it is possible to train a simple support vector machine
classifier which maps the vector code for each node of the tree during decoding to the variable
or function it should represent. This is how we obtain the function names and variables in our
Figures like in Figure~\ref{fig:traces}. We provide the source code for training such a
classifier as part of our code distribution at \url{https://gitlab.com/bpaassen/ast2vec/}.

\paragraph{Training:} Our training procedure is a variation of stochastic gradient descent.
In each training epoch, we randomly sample a mini-batch of
$N = 32$ computer programs from our dataset, compute the abstract syntax trees $\tre x_i$ for
each of them, then encode them as vectors $\Enc(\tre x_i)$ using the current encoder, add Gaussian
noise $\vec \epsilon_i$, and then compute the probabilities
$p_\Dec\big( \tre x_i \big| \Enc(\tre x_i) + \vec \epsilon_i \big)$ as described above.
With these probabilities we compute the loss in Equation~\ref{eq:vae}. Note that this loss
is differentiable with respect to all our neural network parameters. Accordingly, we can use
the automatic differentiation mechanisms of pyTorch \citep{pytorch2019} to compute the gradients of
the loss with respect to all parameters and then perform an Adam update step \citep{Adam2015}.
We set the dimensionality $n$ to $256$, the smoothness parameter $\beta$ to $10^{-3}$, and the
learning rate to $10^{-3}$.

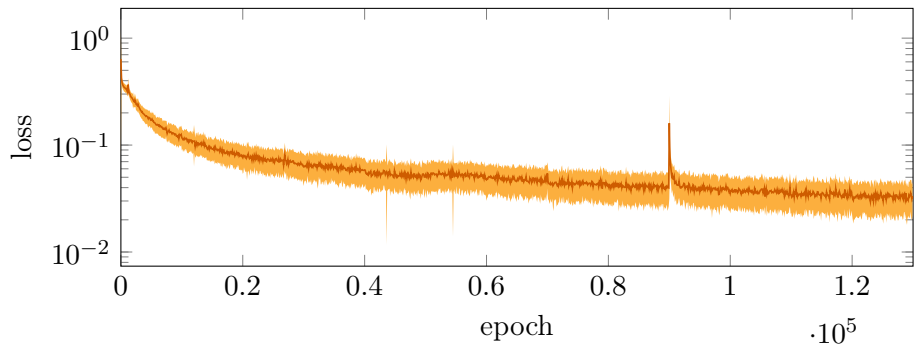
\begin{figure}
\begin{center}
\begin{tikzpicture}
\begin{semilogyaxis}[width=12cm, height=5cm, xlabel={epoch}, ylabel={loss}, xmin=0, xmax=130000]
\addplot[draw=none, name path=hi] table[x=epoch,y=hi] {learning_curve.csv};
\addplot[draw=none, name path=lo] table[x=epoch,y=lo] {learning_curve.csv};
\addplot[orange1] fill between[of=lo and hi];

\addplot[semithick, orange3] table[x=epoch,y=mean] {learning_curve.csv};
\end{semilogyaxis}
\end{tikzpicture}
\end{center}
\caption{The learning curve for our recursive tree grammar autoencoder trained on
$448,992$ Python programs recorded in the 2018 NCSS beginners challenge of groklearning.
The dark orange curve shows the loss of Equation~\ref{eq:vae} over the number of training epochs.
For better readability we show the mean over 100 epochs, enveloped by the standard
deviation (light orange). Note that the plot is in log scaling, indicating little change
toward the end.}
\label{fig:loss}
\end{figure}

We trained ast2vec on $448,992$ Python programs recorded as part of the
the National Computer Science School (NCSS)\footnote{\url{https://ncss.edu.au}}, an Australian
educational outreach programme. The course was delivered by the Grok Learning
platform\footnote{\url{https://groklearning.com/challenge/}}. Each offering was available to
mostly Australian school children in Years 5--10, with curriculum-aligned educational slides
and sets of exercises released each week for five weeks.
For training ast2vec, we used all programs that students tried to run in the 2018 'beginners' challenge.
After compilation, we were left with $86,991$ unique abstract syntax trees.
We performed training for $130,000$ epochs (each epoch with a mini-batch of $32$ trees),
which corresponds to roughly ten epochs per program ($32 \cdot 130,000 = 4,160,000$).
By inspecting the learning curve of the neural net, we notice that the loss has almost
converged (refer to Figure~\ref{fig:loss}).
All training was performed on a consumer-grade laptop with a 2017 Intel core i7
CPU\footnote{Contrary to other neural networks, recursive neural nets can not be trained on
GPUs because the computational graph is unique for each tree.}
and took roughly one week of real time.

We emphasize again that we do not expect teachers to aggregate a similarly huge dataset to
train their own neural net. Instead, we propose to use the pre-trained ast2vec model as a
general-purpose component without retraining (refer to Section~\ref{sec:tutorial}). In the
following sections, we explain two example techniques that are possible thanks to the
vectorial representation achieved by ast2vec.

\subsection{Progress-Variance Projections} \label{sec:analysis_1_construction}

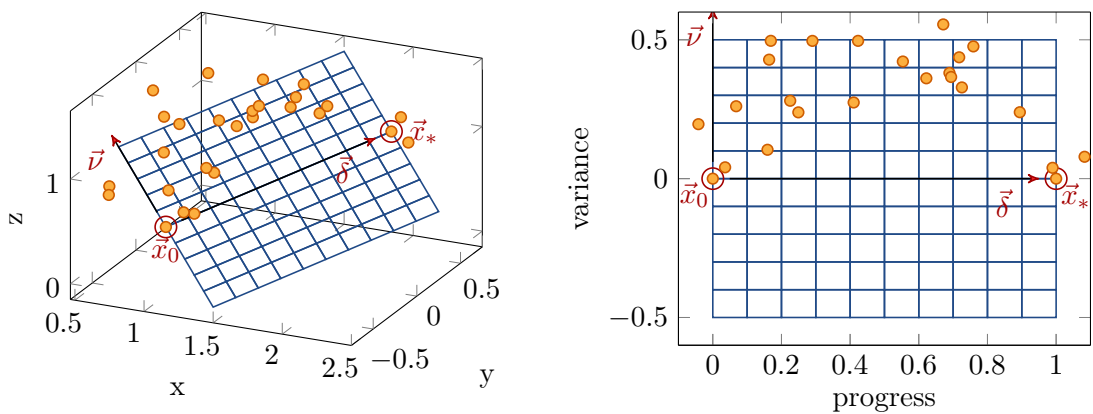
\begin{figure}
\begin{center}
\begin{tikzpicture}
\begin{scope}[shift={(0,0)}]
\begin{axis}[width=7cm, height=6cm, xlabel={x}, ylabel={y}, zlabel={z}]
\addplot3[no marks, semithick, mesh,mesh/rows=11,mesh/ordering=y varies, draw=skyblue3]
table[x=x0, y=x1, z=x2, col sep=tab] {progress_pca_grid_example.csv};
\addplot3[only marks, mark=*, semithick, p1]
table[x=x0, y=x1, z=x2, col sep=tab] {progress_pca_example.csv};

\node[node, draw=scarletred3] (origin) at (axis cs:1,-0.5,0.5) {};
\node[node, draw=scarletred3] (goal)   at (axis cs:2,+0.5,1.) {};

\node[below, scarletred3] at (origin.center) {$\vec x_0$};
\node[right, outer sep=0.1cm, scarletred3] at (goal.center) {$\vec x_*$};

\node (nuend) at (axis cs:0.333,-0.167,1.167) {};

\path[edge, draw=scarletred3, fill=scarletred1]
(origin) edge node[below left, scarletred3, pos=0.9] {$\vec \delta$} (goal)
(origin) edge node[below left, scarletred3, pos=0.9] {$\vec \nu$} (nuend);
\end{axis}
\end{scope}

\begin{scope}[shift={(8,0)}]
\begin{axis}[view={0}{90}, width=7cm, height=6cm, xlabel={progress}, ylabel={variance},
xmin=-0.1,xmax=1.1,ymin=-0.6,ymax=+0.6]
\addplot3[no marks, semithick, mesh,mesh/rows=11,mesh/ordering=y varies, draw=skyblue3]
table[x=y0, y=y1, z expr=0, col sep=tab] {progress_pca_grid_example.csv};
\addplot[only marks, mark=*, semithick, p1]
table[x=y0, y=y1, col sep=tab] {progress_pca_example.csv};

\node[node, draw=scarletred3] (origin) at (axis cs:0,0) {};
\node[node, draw=scarletred3] (goal)   at (axis cs:1,0) {};

\node[below left, inner sep=0.04cm, scarletred3] at (origin.center) {$\vec x_0$};
\node[below right, inner sep=0.06cm, scarletred3] at (goal.center) {$\vec x_*$};

\node (nuend) at (axis cs:0,0.67) {};

\path[edge, draw=scarletred3, fill=scarletred1]
(origin) edge node[below, scarletred3, pos=0.87] {$\vec \delta$} (goal)
(origin) edge node[left, scarletred3, pos=0.8] {$\vec \nu$} (nuend);

\end{axis}
\end{scope}
\end{tikzpicture}
\end{center}
\caption{An illustration of the progress-variance plotting technique. Left: The high-dimensional space
(here 3D) where orange points indicate program encodings.
The encoding for the empty program and the reference solution are highlighted in red and
annotated with $\vec x_0$ and $\vec x_*$, respectively.
The x-axis of our progress-variance plot is the vector $\vec \delta = \vec x_* - \vec x_0$.
The y-axis is the orthogonal axis $\vec \nu$ which covers as much variance as possible of
the point distribution. Right: The 2D-dimensional coordinate system with all points projected
down to 2D.}
\label{fig:progress_pca}
\end{figure}

In this section, we present a novel technique for visualizing student programs, the \emph{progress-variance projection}. This technique, which requires a vector encoding as provided by ast2vec, involves mapping program vectors to a two dimensional representation where the first axis captures the progress between empty program and solution and where the second axis captures the variation orthogonal to the progress axis.
In particular, we map our vectors onto the plane spanned by the following two orthogonal vectors:
\begin{enumerate}
\item $\vec \delta = \vec x_* - \vec x_0$, where $\vec x_*$ and $\vec x_0$ are the encodings of the solution and empty program respectively. This is used as the x-axis of the plot and captures progress towards or away from the solution.
\item $\vec \nu$, which is chosen as the unit vector orthogonal to $\vec \delta$ that preserves as much variance in the dataset as possible. Note that this setup is similar to principal component analysis \citep{Pearson1901}, but with the first component fixed to $\vec \delta$.
\end{enumerate}
The full algorithm to obtain the 2D representation of all vectors is shown in Algorithm~\ref{alg:progress_pca}.

\begin{algorithm}
\caption{An algorithm to map a data matrix of program encodings $\bm{X} \in \R^{N \times n}$
to a 2D progress-variance representation $\bm{Y} \in \R^{N \times 2}$ based on
an encoding for the empty program $\vec x_0 \in \R^n$ and an encoding for the reference
solution $\vec x_* \in \R^n$.}
\label{alg:progress_pca}
\begin{algorithmic}[1]
\State $\vec \delta \gets \vec x_* - \vec x_0$.
\State $\hat \delta \gets \vec \delta / \lVert \vec \delta \rVert$.
\State $\bm{\hat X} \gets \bm{X} - \vec x_0$. \Comment{Row-wise subtraction}
\State $\bm{\hat X} \gets \bm{\hat X} - \bm{\hat X} \cdot \hat \delta \cdot \hat \delta^T$.
\Comment{Project $\vec \delta$ out, i.e.\ $\bm{\hat X}$ lives in the orthogonal space to $\delta$}
\State $\bm{C} \gets \bm{\hat X}^T \cdot \bm{\hat X}$. \Comment{Quasi Covariance matrix}
\State $\vec \nu \gets$ eigenvector with largest eigenvalue of $\bm{C}$.
\State $\bm{Y} \gets (\bm{X} - \vec x_0) / \lVert \vec \delta \rVert \cdot (\hat \delta, \vec \nu)$. \Comment{Map to 2D}
\State \Return $\bm{Y}$.
\end{algorithmic}
\end{algorithm}

Once $\vec \delta$ and $\vec \nu$ are known,
we can translate any high-dimensional vector $\vec x \in \R^n$ to a 2D version $\vec y$ and back via the following equations:

\begin{align}
\vec y &= \begin{pmatrix}
\hat{\delta} & \vec \nu\\
\end{pmatrix}^T \cdot (\vec x - \vec x_0) \  / \  \lVert \vec \delta \rVert & \text{(high to low projection)}\label{eq:hi_to_lo} \\
\vec{x}' &= \begin{pmatrix}
\hat{\delta} & \vec \nu\\
\end{pmatrix} \cdot \vec y \ \times \  \lVert \vec \delta \rVert + \vec x_0, &\text{(low to high embedding)}\label{eq:lo_to_hi}
\end{align}
where $\hat{\delta} = \vec \delta / \lVert \vec \delta \rVert$ is the unit vector parallel to $\vec \delta$ and $\begin{pmatrix}
\hat{\delta} & \vec \nu\\
\end{pmatrix}$ is the matrix with $\hat{\delta}$ and $\vec \nu$ as columns.

In general, $\vec{x}'$ will not be equal to the original $\vec x$ because the 2D space can not
preserve all information in the $n = 256$ dimensions. However, our special construction
ensures that the empty program $\vec x_0$ corresponds exactly to the origin $(0,0)$ and
that the reference solution $\vec x_*$ corresponds exactly to the point $(1,0)$, which can be checked by substituting $\vec x_0$ and $\vec x_*$ into the equations above.
In other words, the $x$-axis represents linear progress in the coding space toward the goal,
and the $y$-axis represents variance orthogonal to progress. An example for the geometric
construction of our progress-variance plot is shown in Figure~\ref{fig:progress_pca}.
In this example, we project a three-dimensional dataset down to 2D.

\subsection{Learning Dynamical Systems}
\label{sec:analysis_2_construction}

In this section, we describe how to learn a linear dynamical system that captures the motion of students
in example data. We note that we are not the first to learn (linear) dynamical systems from
data \citep{Campi1998,Hazan2017}. However, to our knowledge, we are the first to apply dynamical
systems learning for educational datamining purposes; and the system we propose here is particularly simple to learn.

In particular, we consider a linear dynamical system $f$, which predicts the next
step of a student at location $\vec x$ as follows.
\begin{equation}
f(\vec x) = \vec x + \bm{W} \cdot (\vec x_* - \vec x), \label{eq:dynsys}
\end{equation}
where $\vec x_*$ is the reference solution to the task
and $\bm{W}$ is a matrix of parameters to be learned.

This form of the dynamical
system is carefully chosen to ensure two desirable properties.
First, the reference solution is a guaranteed fixed point of our system,
i.e.\ we obtain $f(\vec x_*) = \vec x_*$. Indeed, we can prove that this system has
the reference solution $\vec x_*$ as unique stable attractor if $\bm{W}$ has
sufficiently small eigenvalues (refer to Theorem~\ref{thm:stability} in the appendix
for details). This is desirable because it guarantees that the default behavior of our
system is to move towards the correct solution.

Second, we can \emph{learn} a matrix $\bm{W}$ that best describes student motion
via simple linear regression. In particular, let $\vec x_1, \ldots, \vec x_T$
be a sequence of programs submitted by a student in their vector encoding provided by ast2vec.
Then, we wish to find the matrix $\bm{W}$ which
best captures the dynamics in the sense that $f(\vec x_t)$ should be as close
as possible to $\vec x_{t+1}$ for all $t$. More formally, we obtain the following
minimization problem.
\begin{equation}
\min_{\bm{W}} \quad \sum_{t=1}^{T-1} \lVert f(\vec x_t) - \vec x_{t+1}\rVert^2 + \lambda \cdot \lVert \bm{W} \rVert^2_\mathcal{F}, \label{eq:dynsys_learning}
\end{equation}
where $\lVert \bm{W} \rVert_\mathcal{F}$ is the Frobenius norm of the matrix $\bm{W}$
and where $\lambda > 0$ is a parameter that can be increased to ensure that the reference solution remains an attractor. This problem has the closed-form solution
\begin{equation}
\bm{W} = \big(\bm{X}_{t+1}-\bm{X}_t \big)^T \cdot \bm{X}_t \cdot \big(\bm{X}_t^T \cdot \bm{X}_t + \lambda \cdot \bm{I}\big)^{-1}, \label{eq:linreg}
\end{equation}
where $\bm{X}_t = (\vec x_* - \vec x_1, \ldots,
\vec x_* - \vec x_{T-1})^T \in \R^{T - 1 \times n}$ is the concatenation
of all vectors in the trace up to the last one and
$\bm{X}_{t+1} = (\vec x_* - \vec x_2, \ldots,
\vec x_* - \vec x_T)^T \in \R^{T-1 \times n}$ is the concatenation of all successors. Refer to Theorem~\ref{thm:linreg} in the appendix for a proof.

As a side note, we wish to highlight that this technique can readily be extended to multiple
reference solutions by replacing $\vec x_*$ in Equations~\ref{eq:dynsys} and~\ref{eq:linreg}
with the respective closest correct solution to the student's current state. In other words,
we partition the space of programs according to several basins of attraction, one per correct
solution. The setup and training remain the same, otherwise.

This concludes our description of methods. In the next section, we evaluate these
methods on large-scale datasets.

\section{Evaluation} \label{sec:evaluation}

In this section, we evaluate the ast2vec model on
two large-scale anonymised datasets of introductory programming, namely the 2018 and 2019 beginners 
challenge by the National Computer Science School (NCSS)\footnote{\url{https://ncss.edu.au}}, an Australian educational
outreach programme. The courses were delivered by the Grok Learning platform\footnote{\url{https://groklearning.com/challenge/}}. Each offering was available to (mostly) Australian school children in Years 5--10, with curriculum-aligned educational slides and sets of exercises released each week for five weeks. Students received a score for successfully completing each exercise, with the score available starting at 10 points, reducing by one point every 5 incorrect submissions, to a minimum of 5 points. In both datasets, we consider only submissions, i.e.\ programs that students
deliberately submitted for evaluation against test cases.

The 2018 dataset contains data of $12,141$ students working on $26$ different
programming tasks, yielding $148,658$ compileable programs.
This is also part of the data on which ast2vec was trained.
The 2019 dataset contains data of $10,558$ students working on $24$ problems, yielding
$194,797$ compileable programs overall.

For our first analysis, we further broaden our scope and include a third dataset with a slightly different course format, in particular the Australian Computing Academy digital technologies (DT) chatbot project\footnote{\url{https://aca.edu.au/resources/python-chatbot/}} which consists of 63 problems and teaches students the skills to program a simple chatbot, and is also delivered via the Grok Learning platform. Our dataset includes the data of $27,480$ students enrolled between May 2017 and August 2020, yielding $1,343,608$ compilable programs.

We first check how well ast2vec is able to correctly autoencode trees in all three datasets and
then go on to check its utility for prediction. We close the section by inspecting the coding
space in more detail for the example dataset from Section~\ref{sec:tutorial}.

\subsection{Autoencoding error}

Our first evaluation concerns the ability of ast2vec to maintain information in its encoding.
In particular, we evaluate the autoencoding error, i.e.\ we iterate over all compileable
programs in the dataset, compile them to an abstract syntax tree (AST), use ast2vec to convert
the AST into a vector, decode the vector back into an AST, and compute the tree edit distance \citep{Zhang1989}
to the original AST.

\begin{figure}
\begin{center}
\begin{tikzpicture}
\begin{axis}[width=14cm,height=5cm,
axis y line*=right,
axis line style=skyblue3,
every axis label/.append style ={skyblue3},
every tick label/.append style ={skyblue3},
axis x line=none,
xmin=0,xmax=70,
ymin=0,ymax=100,ylabel={coverage [\%]},
]
\addplot[skyblue3, thick, dashed 
]
table[x=size,y=cum_num,col sep=tab] {beginners_2018_errors.csv};
\end{axis}
\begin{axis}[width=14cm,height=5cm,
xmin=0,xmax=70,xlabel={tree size},
ymin=0,ymax=50,ylabel={error (TED)},
axis y line*=left,
title={NCSS beginners 2018 data}]
\addplot[draw=none, name path=hi]
table[x=size,y expr=\thisrow{mean}+\thisrow{std},col sep=tab] {beginners_2018_errors.csv};
\addplot[draw=none, name path=lo]
table[x=size,y expr=\thisrow{mean}-\thisrow{std},col sep=tab] {beginners_2018_errors.csv};
\addplot[orange1, opacity=0.3] fill between[of=lo and hi];

\addplot[orange3, thick]
table[x=size,y=mean,col sep=tab] {beginners_2018_errors.csv};
\addplot[aluminium6, thick]
table[x=size,y=median,col sep=tab] {beginners_2018_errors.csv};
\end{axis}
\end{tikzpicture}
\begin{tikzpicture}
\begin{axis}[width=14cm,height=5cm,
axis y line*=right,
axis line style=skyblue3,
every axis label/.append style ={skyblue3},
every tick label/.append style ={skyblue3},
axis x line=none,
xmin=0,xmax=70,
ymin=0,ymax=100,ylabel={coverage [\%]},
]
\addplot[skyblue3, semithick, dashed, thick]
table[x=size, y = cum_num, col sep=tab] {beginners_2019_errors.csv};
\end{axis}
\begin{axis}[
width=14cm,
height=5cm,
xmin=0,xmax=70,xlabel={tree size},
ymin=0,ymax=50,ylabel={error (TED)},
axis y line*=left,
title={NCSS beginners 2019 data}]
\addplot[draw=none, name path=hi]
table[x=size,y expr=\thisrow{mean}+\thisrow{std},col sep=tab] {beginners_2019_errors.csv};
\addplot[draw=none, name path=lo]
table[x=size,y expr=\thisrow{mean}-\thisrow{std},col sep=tab] {beginners_2019_errors.csv};
\addplot[orange1, opacity=0.3] fill between[of=lo and hi];

\addplot[orange3, semithick]
table[x=size,y=mean,col sep=tab] {beginners_2019_errors.csv};
\addplot[aluminium6, semithick]
table[x=size,y=median,col sep=tab] {beginners_2019_errors.csv};
\end{axis}
\end{tikzpicture}
\begin{tikzpicture}
\begin{axis}[width=14cm,height=5cm,
axis y line*=right,
axis line style=skyblue3,
every axis label/.append style ={skyblue3},
every tick label/.append style ={skyblue3},
axis x line=none,
xmin=0,xmax=70,
ymin=0,ymax = 100,ylabel={coverage [\%]},
]
\addplot[skyblue3, thick, dashed]
table[x=size,y=cum_num,col sep=tab] {chatbot_error.csv};
\end{axis}
\begin{axis}[width=14cm,height=5cm,
xmin=0,xmax=70,xlabel={tree size},
ymin=0,ymax=50,ylabel={error (TED)},
axis y line*=left,
title={DT Chatbot data}]
\addplot[draw=none, name path=hi]
table[x=size,y expr=\thisrow{mean}+\thisrow{std},col sep=tab] {chatbot_error.csv};
\addplot[draw=none, name path=lo]
table[x=size,y expr=\thisrow{mean}-\thisrow{std},col sep=tab] {chatbot_error.csv};
\addplot[orange1, opacity=0.3] fill between[of=lo and hi];

\addplot[orange3, thick]
table[x=size,y=mean,col sep=tab] {chatbot_error.csv};
\addplot[aluminium6, thick]
table[x=size,y=median,col sep=tab] {chatbot_error.csv};
\end{axis}
\end{tikzpicture}
\end{center}
\caption{The autoencoding error as measured by the tree edit distance
for the NCSS 2018 beginners challenge (top), NCSS 2019 beginner challenge (middle) and DT chatbot course (bottom). The error is plotted here versus tree
size. The orange line marks the mean, the black line the median error. The
orange region is the standard deviation around the man. Additionally, the blue
line indicates how many trees in the dataset have a size up to $x$.}
\label{fig:autoencoding_errors}
\end{figure}
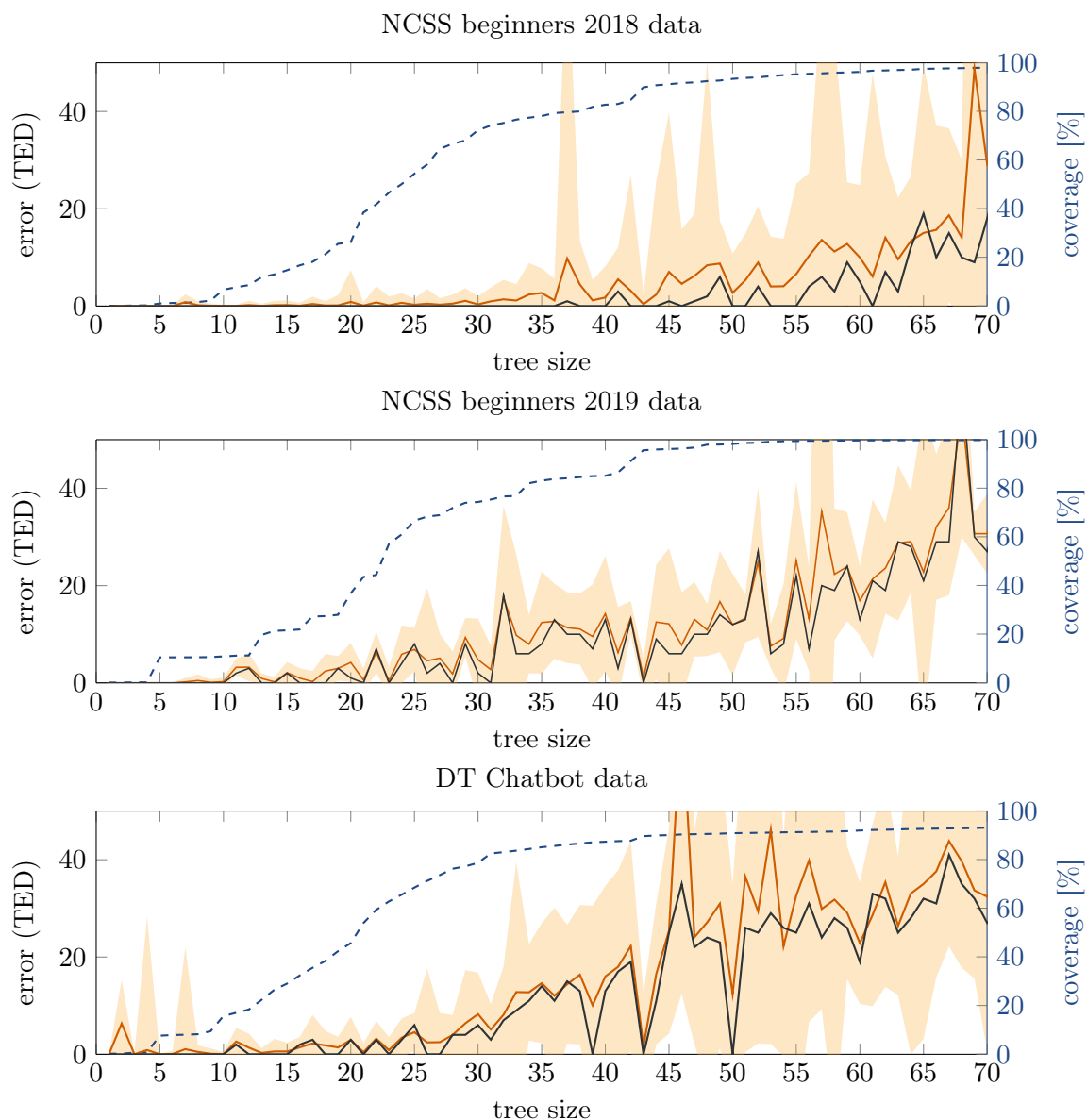

Figure~\ref{fig:autoencoding_errors} shows the autoencoding error 
for trees of different sizes in the three datasets.
The black line shows the median autoencoding error, whereas the orange line shows the average,
with the orange region indicating one standard deviation. The blue dashed line
indicates the cumulative distribution, i.e.\ the fraction of trees in the data that have at most $x$ nodes.

With respect to the 2018 data, we observe that the median is generally
lower than the mean, indicating that the distribution has a long tail of few trees with
higher autoencoding error, whereas most trees have low autoencoding error. Indeed, up to tree
size $36$ the median autoencoding error is zero, and $78.92\%$ of all programs in the dataset
are smaller than $37$ nodes, thus covering a sizeable portion of the dataset.
The overall average autoencoding error across the dataset is $1.86$
and the average tree size is $27.98$.

With respect to the 2019 dataset, we observe that median and average are more closely aligned, indicating more
symmetric distributions of errors. Additionally, we notice that the autoencoding
error grows more quickly compared to the 2018 data, which indicates that ast2vec is slightly worse in autoencoding programs
outside its training data compared to programs inside its training data. 
We obtain an average of $4.10$ compared to an average tree size of $24.28$.

Similarly, for the chatbot data, we observe the average overtaking the median and that the
error is generally larger compared to the 2018 data. A specific property of the
chatbot dataset is that there is a long tail of large programs which correspond to the full
chatbot. If we include these in the analysis, we obtain an average autoencoding error of
$7.86$ and an average tree size of $27.40$.
If we only consider trees up to size $85$, covering 95\% of all trees in the dataset,
we obtain an average autoencoding error of $4.41$ and an average tree size of $22.49$.

Overall, we observe that ast2vec performs better on smaller trees and better on trees in the training data. This is not
surprising but should be taken into account when applying ast2vec to new datasets. One may also be able to further reduce the autoencoding error by adjusting the ast2vec architecture, e.g.\ by incorporating more recurrent nets such as Tree LSTMs \citep{Chen2018,Dai2018,Tai2015}.

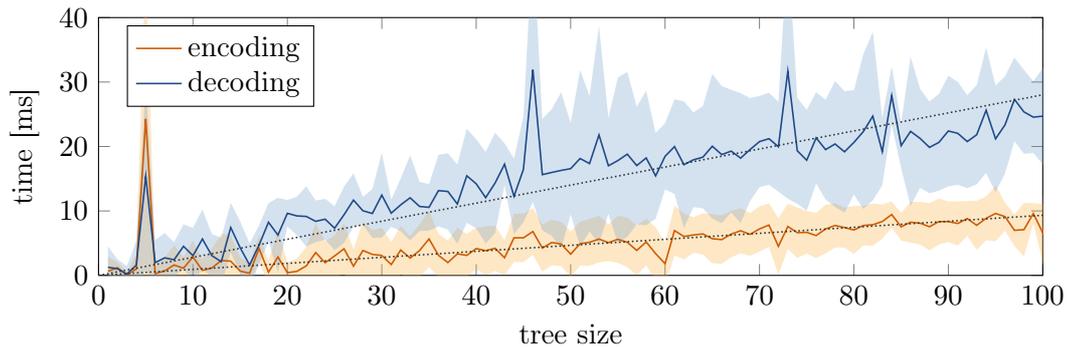
\begin{figure}
\begin{center}
\begin{tikzpicture}
\begin{axis}[width=14cm,height=5cm,
xmin=0,xmax=100,xlabel={tree size},
ymin=0,ymax=40,ylabel={time [ms]},
legend pos={north west}, legend cell align={left}]
\addplot[draw=none, name path=hi, forget plot]
table[x=size,y expr=1000*(\thisrow{mean}+\thisrow{std}),col sep=tab] {chatbot_enc_time.csv};
\addplot[draw=none, name path=lo, forget plot]
table[x=size,y expr=1000*(\thisrow{mean}-\thisrow{std}),col sep=tab] {chatbot_enc_time.csv};
\addplot[orange1, opacity=0.3, forget plot] fill between[of=lo and hi];
\addplot[orange3, semithick]
table[x=size,y expr=1000*\thisrow{mean},col sep=tab] {chatbot_enc_time.csv};
\addlegendentry{encoding}
\addplot[aluminium6, semithick, densely dotted, domain=0:100, samples=11, forget plot]
{0.09328104*x};

\addplot[draw=none, name path=hi, forget plot]
table[x=size,y expr=1000*(\thisrow{mean}+\thisrow{std}),col sep=tab] {chatbot_dec_time.csv};
\addplot[draw=none, name path=lo, forget plot]
table[x=size,y expr=1000*(\thisrow{mean}-\thisrow{std}),col sep=tab] {chatbot_dec_time.csv};
\addplot[skyblue1, opacity=0.3, forget plot] fill between[of=lo and hi];
\addplot[skyblue3, semithick]
table[x=size,y expr=1000*\thisrow{mean},col sep=tab] {chatbot_dec_time.csv};
\addlegendentry{decoding}
\addplot[aluminium6, semithick, densely dotted, domain=0:100, samples=11, forget plot]
{0.28016169*x};
\end{axis}
\end{tikzpicture}
\end{center}
\caption{The time needed for encoding (orange) and decoding (blue)
trees of different sizes for the DT chatbot course. The thick lines mark the means whereas shaded regions
indicate one standard deviation. The dotted lines indicate the best linear fit.}
\label{fig:chatbot_times}
\end{figure}

Figure~\ref{fig:chatbot_times} shows the time needed to encode (orange)
and decode (blue) a tree from the chatbot dataset of a given size. 
As the plot indicates, the empiric time complexity for both operations is roughly
linear in the tree size, which corresponds to the theoretical findings of \citet{Paassen2021}.
Using a linear regression without intercept, we find that encoding requires roughly $0.9$
millisecond per ten tree nodes, whereas decoding requires roughly $2.8$ milliseconds per
ten tree nodes. Given that decoding involves more operations (both element choice and
vector operations), this difference in runtime is to be expected. Fortunately, both
operations remain fast even for relatively large trees.

\subsection{Next-Step Prediction}
In the previous section, we evaluated how well ast2vec could translate from programs to vectors and back, finding that it did well on the majority of beginner programs. In this section, we investigate the dynamical systems analysis suggested in Sections~\ref{sec:dynamics} and~\ref{sec:analysis_2_construction}. In particular, we compare its ability to predict the next step of a student based on example data from other students.

Recall that our dynamical systems model is quite simple: It encodes the student's current abstract syntax tree as a vector, computes the difference in the encoding space to the most common correct solution, applies a linear transformation to that difference, adds the result to the student's current position, and decodes the resulting vector to achieve the next-step-prediction. Given this simplicity, our research objective in this section is not to provide the best possible next-step prediction. Rather, we wish to investigate how well a very simple model can perform just by virtue of using the continuous vector space of ast2vec.

By contrast, we compare to the following reference models from the literature:
\begin{enumerate}
\item the simple identity, which predicts the next step to be the same as the current one and is included as a baseline \citep{Hyndman2006},
\item one-nearest-neighbor prediction (1NN). This involves searching the training data for the closest tree based on tree edit distance and predicting its successor, which is similar to the scheme suggested by \citet{Gross2015}, and
\item the continuous hint factory \cite[CHF]{Paassen2018JEDM}, which is also based on the tree edit distance but instead involves Gaussian process regression and heuristic-driven reconstruction techniques to predict next steps.
\end{enumerate}

Both 1NN as well as CHF are nonlinear predictors with a much higher representational capacity, such that we would expect them to be more accurate \citep{Paassen2017}. We investigate this in the following section.

\subsubsection{Predictive error}

We evaluate the error in next-step prediction on two different datasets: the first was the 2018 NCSS challenge data, which was used to train ast2vec, and the second was the 2019 NCSS challenge data, which contained programs unseen by ast2vec. We considered each learning task in both challenges separately. For each task, we partitioned the student data into 10 folds and performed 10-fold cross-validation, i.e.\ using nine folds as training data and one fold as evaluation data. Additionally, we subsampled the training data to only contain the data of 30 student in order to simulate a classroom-sized training dataset.

\begin{figure}
\begin{center}
\begin{tikzpicture}
\begin{axis}[
    width=13.5cm,
    height=5cm,
    xlabel={task number}, 
    ylabel={prediction error (TED)}, 
    xmin=1,
    xmax=26,
    ymin=0,
    legend pos={outer north east},
    legend style={cells={align=left,anchor=west}},
    title={2018 NCSS challenge}
]
\addplot[aluminium6, semithick] table[x expr=\lineno, y=baseline_mean] {prediction_errors_30_students.csv};
\addlegendentry{baseline}
\addplot[mark=*, every mark/.append style={fill=skyblue1}, skyblue3, semithick, opacity=0.8] table[x expr=\lineno, y=1nn_mean] {prediction_errors_30_students.csv};
\addlegendentry{1NN}
\addplot[mark=square*, every mark/.append style={fill=orange1}, orange3, semithick, opacity=0.8] table[x expr=\lineno, y=chf_mean] {prediction_errors_30_students.csv};
\addlegendentry{CHF}
\addplot[mark=triangle*, every mark/.append style={fill=orange1}, scarletred3, semithick, opacity=0.8] table[x expr=\lineno, y=linear_mean] {prediction_errors_30_students.csv};
\addlegendentry{ast2vec\\  + linear};
\end{axis}
\end{tikzpicture}
\begin{tikzpicture}
\begin{axis}[width=13.5cm,height=5cm,
xlabel={task number}, ylabel={prediction error (TED)}, xmin=1,xmax=24,ymin=0,
legend pos={outer north east}, legend style={cells={align=left,anchor=west}},
title={2019 NCSS challenge}]
\addplot[aluminium6, semithick] table[x expr=\lineno, y=baseline_mean] {prediction_errors_30_students_2019.csv};
\addlegendentry{baseline}
\addplot[mark=*, every mark/.append style={fill=skyblue1}, skyblue3, semithick, opacity=0.8] table[x expr=\lineno, y=1nn_mean] {prediction_errors_30_students_2019.csv};
\addlegendentry{1NN}
\addplot[mark=square*, every mark/.append style={fill=orange1}, orange3, semithick, opacity=0.8] table[x expr=\lineno, y=chf_mean] {prediction_errors_30_students_2019.csv};
\addlegendentry{CHF}
\addplot[mark=triangle*, every mark/.append style={fill=orange1}, scarletred3, semithick, opacity=0.8] table[x expr=\lineno, y=linear_mean] {prediction_errors_30_students_2019.csv};
\addlegendentry{ast2vec\\  + linear};
\end{axis}
\end{tikzpicture}
\end{center}
\caption{The average root mean square prediction error (in terms of tree edit distance) across the entire beginners 2018 (top) and 2019 (bottom) curriculum for
different prediction methods. The x-axis uses the same order of problems as students worked
on them.}
\label{fig:prediction_error_2018_2019}
\end{figure}
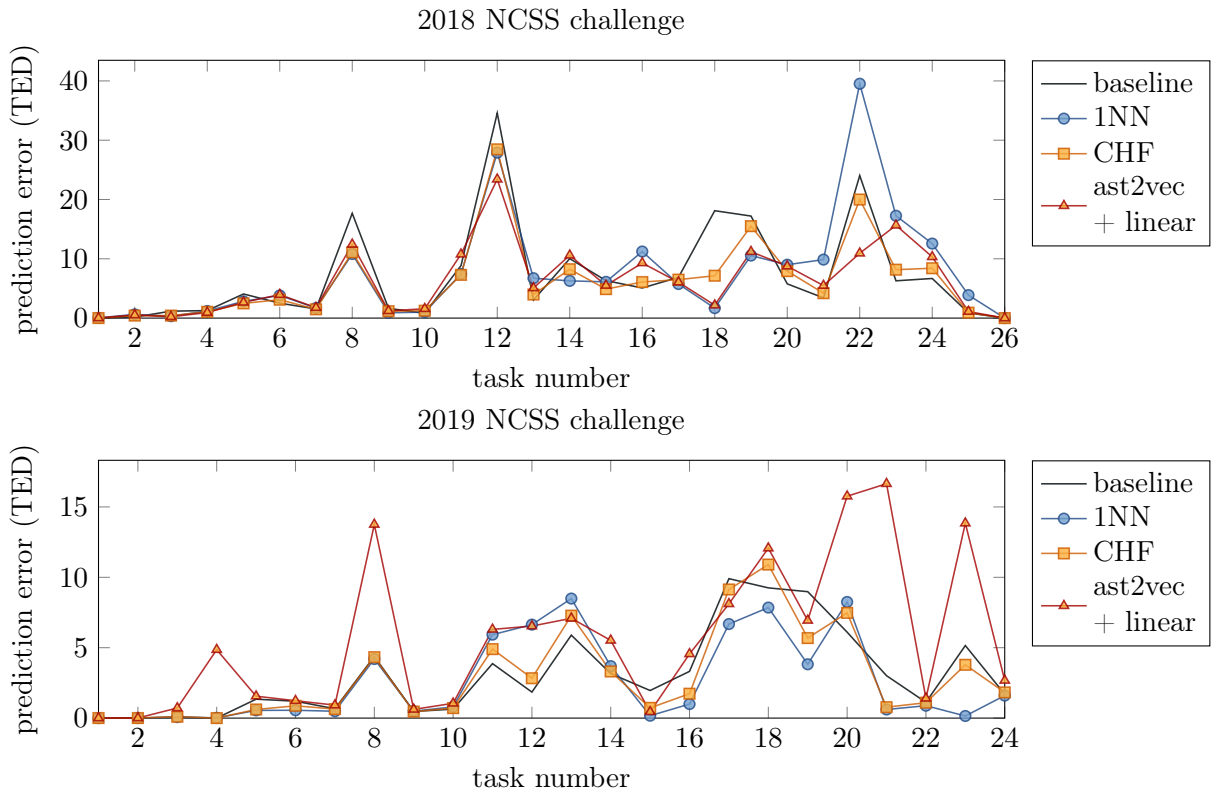

Figure~\ref{fig:prediction_error_2018_2019} 
shows the average root mean square prediction error in terms of tree edit distance on the 2018 and 2019 data. In general, our simple linear model performed comparably to the other models, with the performance being particularly close for the 2018 data. This difference between the datasets is to be expected given that ast2vec has lower autoencoding error on the 2018 data set.

With respect to the 2019 data, our model still performed reasonably well for most exercises, though there were some cases where the error was noticeably higher (i.e.\ tasks 4, 8, 20, 21 and 23). The error on tasks 4, 20, 21 and 23 can be explained by the fact that these tasks involved larger programs (task 4 was an early exercise, but it involved a large program scaffold that students needed to modify). Based on the analysis in the previous section, larger programs tend to have higher autoencoding error, which we would expect to affect the prediction error. Task 8, however, is interesting because it involves relatively small programs. On closer inspection, this task has many different possible solutions, so the error could be related to the simple design of our model (specifically, that it only makes predictions towards the most common solution). All-in-all, this suggests the simple model performs comparably to the others, and only performs worse in a small number of specific and explainable cases.

We note in passing that the naive baseline of predicting the current step performs also quite well in both datasets. This may seem surprising but is a usual finding in forecasting settings \citep{Hyndman2006,Paassen2017}. In particular, a low baseline error merely indicates that students tend to change their program only a little in each step, which is to be expected. Importantly, though, the baseline can not be used in an actual hint system, where one would utilize the difference between a student's current state and the predicted next state to generate hints \citep{Rivers2017,Paassen2018JEDM}. For the baseline, there is no difference, and hence no hints can be generated.

\subsubsection{Runtime}

\begin{figure}
\begin{center}
\begin{tikzpicture}
\begin{semilogyaxis}[width=13.5cm,height=5cm,
xlabel={number of students}, ylabel={prediction time [s]}, xmin=10,xmax=500,ymin=0.01,ymax=5,
legend style={cells={align=left,anchor=west}},
legend pos={outer north east}]
\addplot[skyblue3, semithick, mark=*, every mark/.append style={fill=skyblue1}] table[x=num_students, y=1nn_mean] {predict_times_2019.csv};
\addlegendentry{1NN}
\addplot[orange3, semithick, mark=square*, every mark/.append style={fill=orange1}] table[x=num_students, y=chf_mean] {predict_times_2019.csv};
\addplot[orange3, densely dashed, semithick, mark=square*, every mark/.append style={fill=orange1}, forget plot] table[x=num_students, y=chf_mean] {train_times_2019.csv};
\addlegendentry{CHF}
\addplot[scarletred3, semithick, mark=triangle*, every mark/.append style={fill=scarletred1}] table[x=num_students, y=linear_mean] {predict_times_2019.csv};
\addplot[scarletred1, densely dashed, semithick, mark=triangle*, every mark/.append style={fill=scarletred1}, forget plot] table[x=num_students, y=linear_mean] {train_times_2019.csv};
\addlegendentry{ast2vec\\ + linear}
\end{semilogyaxis}
\end{tikzpicture}
\end{center}
\caption{The average training (dashed lines) and prediction time (solid lines) for
different training set sizes for the 2019 dataset. Prediction times are measured
as the accumulated time to provide predictions for an entire student trace.
Note that the y axis is in log scale.}
\label{fig:predict_times}
\end{figure}
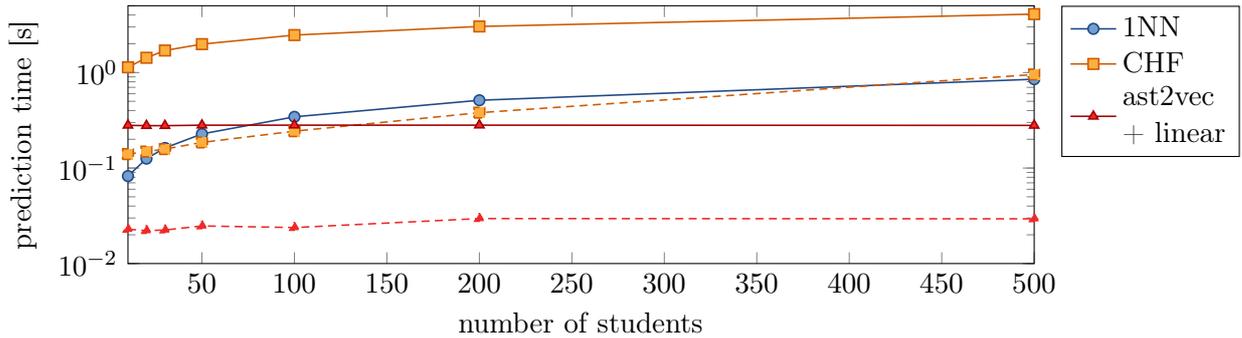

Another interesting aspect is training and prediction time. While training is very fast
across all methods (1NN does not even need training), the prediction time of neighborhood-based
methods like 1NN and CHF scales with the size of the training dataset, whereas the prediction
time for our ast2vec+linear scheme remains constant. Figure~\ref{fig:predict_times} displays
the average training and prediction times for all methods and varying training dataset sizes
on the 2019 dataset. As we can see, our proposed scheme always needs about $300$ms to make
all predictions for a student trace (including encoding and decoding times), whereas 1NN
exceeds this time starting from $\approx 80$ students and CHF is slower across time scales,
requiring more than a second even for small training data sizes. This is an important consideration in large-scale educational contexts, where many predictions may need to be made very quickly.

\subsubsection{Qualitative Comparison}

Beyond the quantitative comparison in terms of prediction error and runtime, there are important qualitative differences between our proposed dynamical systems and previous, neighborhood-based schemes.
As noted in Section \ref{sec:analysis_2_construction}, one important property of our simple model is that the predictions are mathematically guaranteed to lead to a solution, and the linearity of the model means this is done smoothly. By contrast, neither 1NN nor CHF can formally guarantee convergence to a correct solution if one follows the predictions. Additionally, the predictions of 1NN are discontinuous, i.e.\ they change suddenly if the student enters a different neighborhood. Overall, a smooth and guaranteed motion towards a correct solution could be valuable for designing a next-step hint system where the generated hints should be both directed to the desired target and intuitive \citep{McBroom2019,Paassen2018JEDM,Rivers2017}.

Furthermore, a dynamical system based on ast2vec only has to solve the problem of describing student motion in the space of programs for a particular task. Constructing the space of programs in the first place is already solved by ast2vec. By contrast, a neighborhood-based model must solve both the representation \emph{and} the learning problem anew for each learning task.

Additionally, any neighborhood-based model must store the entire training dataset at all times and recall it for every prediction, whereas the ast2vec+linear model merely needs to store and recover the predictive parameters, which includes the ast2vec parameters, and for each learning task the vector encoding of the correct solution and the matrix $\bm{W}$. This is not only more time-efficient (as we saw above) but also space efficient as each new learning task only requires an additional $n+1 \times n = 65,792$ floating point numbers to be stored, which is roughly half a megabyte.

Finally, if additional predictive accuracy is desired, one can improve the predictive model by replacing the linear prediction with a nonlinear prediction (e.g.\ a small neural network), whereas the predictive accuracy of a neighborhood-based system can only be improved by using a richer distance measure or more training data.

In summary, this evaluation has shown that a simple linear regression model using the vector encodings of ast2vec can perform comparably to neighborhood-based next-step prediction techniques in terms of prediction error, but has other desirable properties that may make it preferable in many settings, such as constant time and space complexity, guaranteed convergence to a correct solution, smoothness, and separation of concerns between representation and prediction. As such, we hope that ast2vec can contribute to improve educational datamining pipelines, e.g.\ for next-step hint predictions in the future.

\subsection{Interpolation in the coding space} \label{sec:analysis_1_variation2}

In the previous sections we have established that ast2vec is generally able to decode vectors
back to fitting syntax trees. However, what remains open is whether the coding space is
\emph{smooth}, i.e.\ whether neighboring points in the coding space correspond to similar
programs. This property is important to ensure that typical analyses like visualization and
dynamical systems from Section~\ref{sec:tutorial} are meaningful. More specifically, if we
look at a visualization of a dataset, we implicitly assume that points close in the plot also
correspond to similar programs. Similarly, when we perform a dynamical system analysis, we
implicitly assume that small movements in the $n = 256$-dimensional coding space also correspond
to small changes in the corresponding program.

In principle, the variational autoencoder framework (refer to Section~\ref{sec:method}) ensures such a smoothness property. In this
section, we wish to validate this for an example. In particular, Figure~\ref{fig:interp}
shows a 2D grid of points which we sample from the progress-variance projection for our
example dataset from Section~\ref{sec:tutorial}. Points that decode to the same syntax tree
receive the same color.
The programs corresponding to regions with at least 10 grid points are shown on the right.

This visualization shows several reassuring properties: First, neighboring points in the grid
tend to decode to the same tree. Second, if two points correspond to the same tree, the
space between them also correspond to the same tree, i.e.\ there are no two points
corresponding to the same tree at completely disparate locations of the grid. Finally, the syntax trees
that are particularly common are also meaningful for the task, in the sense that they actually
occur in student data. Overall, this bolsters our confidence that the encoding space is indeed
smooth.

\pgfplotsset{colormap={rainbow}{[1cm] color(0cm)=(plum2) color(1cm)=(skyblue2) color(2cm)=(chameleon2) color(3cm)=(butter2) color(4cm)=(orange2) color(5cm)=(scarletred2)}}

\begin{figure}
\begin{center}
\begin{tikzpicture}
\begin{axis}[view={0}{90}, xlabel={progress}, ylabel={variance},
enlarge x limits=0.05, enlarge y limits=0.05, width=7cm, height=6cm]
\addplot3+[mark size=4, semithick, mesh,scatter,mesh/rows=11,mesh/ordering=y varies]
table[x=i, y=j, z=tree_index, col sep=tab] {pca_grid.csv};

\node (start) at (axis cs:0.2,-0.3) {};
\node (intermediate) at (axis cs:0.5,0.4) {};
\node (goal) at (axis cs:0.9,0) {};

\end{axis}

\node[fill=white,draw=black] at (start.center) {
\footnotesize{empty program}
};

\begin{scope}[shift={(6.5,-0.55)}]
\node[inner sep=0.1cm, fill=white,draw=black,align=left, below right] (intermediate_prog) at (0,5) {
\begin{lstlisting}
x = input('<string>')
print('<string>')
\end{lstlisting}
};

\node[inner sep=0.1cm, fill=white,draw=black,align=left, below right] (goal_prog) at (0,3) {
\begin{lstlisting}
x = input('<string>')
if x == '<string>':
  print('<string>')
else:
  print('<string>')
\end{lstlisting}
};
\end{scope}

\path[edge]
(intermediate_prog.west) edge[out=180,in=-30] (intermediate)
(goal_prog.west) edge[out=180,in=-45] (goal);

\end{tikzpicture}
\end{center}
\caption{A linear interpolation between the empty program at $(0,0)$
and the correct solution at $(1, 0)$ as well as along the axis
of biggest variation. Color indicates the tree that the grid point decodes to.
The three most common trees are shown in the boxes.}
\label{fig:interp}
\end{figure}
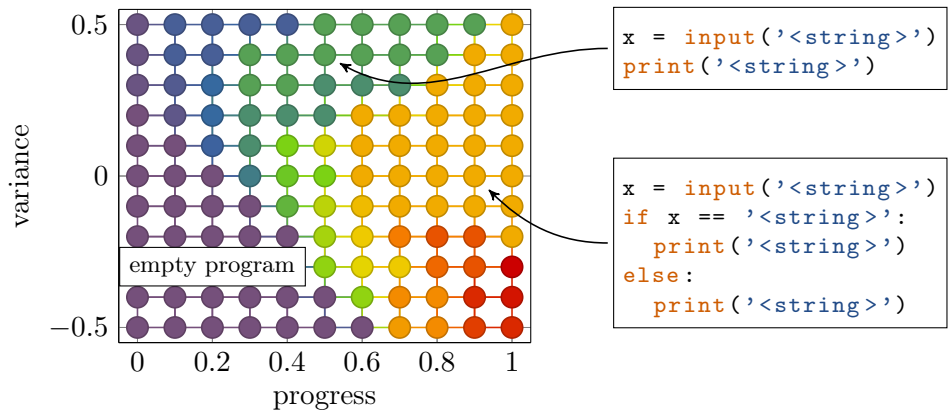

\section{Related Work} \label{sec:related_work}

In this paper, we propose a novel, vectorial encoding for computer programs to
support educational datamining. In doing so, we build on prior work which also proposed
alternative representations for computer programs. In the remainder of this section, we
review these alternative representations and relate them to ast2vec.

\paragraph{Abstract Syntax Trees:} An abstract syntax tree (AST) represents the program code as
a tree of syntactic elements,
which is also the internal representation used by compilers to transform human-readable program
code to machine code \citep{Aho2006}.
For example, Figure~\ref{fig:encoding} (left) shows the abstract syntax
tree for the Python program \lstinline!print('Hello, world!')!.
In most processing pipelines for computer programs, compiling the source code into an AST is the
first step, because it tokenizes the sources code meaningfully, establishes the hierarchical
structure of the program, and separates functionally relevant syntax from functionally irrelevant
syntax, like variable names or comments \citep{Aho2006,McBroom2019}.
Additionally, several augmentations have been suggested to make syntax trees more meaningful.
In particular, \citet{Rivers2012} suggest canonicalization rules to match trees
that are functionally equivalent, such as normalizing the direction of comparison operators,
inlining helper functions, or removing unreachable code. \citet{Gross2014} further suggest
to insert edges in the tree between variable names and their original declaration,
thus augmenting the tree to a graph. Following prior work, we also use the AST representation as
a first step before we apply a neural network to it. Although we do not use them here explicitly,
the canonicalization rules of \citet{Rivers2012} are directly compatible with ast2vec.
A more difficult match are the additional reference edges suggested by \citet{Gross2014}
because these would require neural networks that can deal with full graphs, i.e.\ 
graph neural networks \citep{Kipf2016,Scarselli2009}.

\paragraph{Tree Patterns:} Since many datamining methods can not be applied to trees,
several prior works have suggested to transform the AST to a collection of patterns first.
Such representations are common in the domain of natural language processing, where texts
can be represented as a collection of the words or $n$-grams they contain
\citep{Broder1997}. Such techniques can be extended to trees by representing
a tree by the subtree patterns it contains \citep{Nguyen2014,Zhi2018,Zimmerman2015}.
For example, the syntax tree \enquote{Module(Expr(Call(Name, Str)))} could be represented as
the following collection of subtree patterns of height two: \enquote{Call(Name, Str)},
\enquote{Expr(Call)}, and \enquote{Module(Expr)}. Once this collection is computed, we can
associate all possible subtree patterns with a unique index in the range $1, \ldots, K$ and then
represent a tree by a $K$-dimensional vector, where entry $k$ counts how often the $k$th subtree
pattern occurs in the tree.

If the collection of subtree patterns is meaningful for the programming task at hand, this
representation can be both simple and powerful. For example, if a programming task is about
writing a while loop, it is valuable to know whether a student's program actually contains such
a loop. \citet{Zhi2018} have considered this issue in more detail and considered both
expert-based and data-driven tree patterns as a representation. However, if there is no clear
relation between tree patterns and progress in a task, it may be problematic to use such
pattern as a representation, in particular when all tree patterns are weighted equally.
ast2vec is weakly related to tree patterns because the tree pattern determines which encoding/decoding
functions are called when processing the tree. However, the vector returned by the network does
not simply count tree patterns but, instead, considers the entire tree structure, i.e.\ how the
tree patterns are nested into each other to form the entire tree. More precisely, ast2vec is trained
to generate vector encodings that still contain enough information to recover the original tree,
whereas tree pattern counting usually does not enable us to recover the original tree.

\paragraph{Pairwise distances:} An alternative to an explicit vector representation is
offered by pairwise distances (or similarities). Most prominently, abstract syntax trees can
be represented by their pairwise tree edit distance $d(x, y)$, which is defined as the number
of syntactic elements one has to delete, insert, or re-label to get from tree $x$ to
tree $y$ \citep{Mokbel2013,Paassen2018JEDM,Price2017,Rivers2017,Zhang1989}. Such a pairwise
distance representation is mathematically quite rich. Indeed, one can prove that a distance
representation implicitly assigns vectors to programs and one can facilitate this implicit
representation for distance-based variants of machine learning methods, e.g.\ for visualization,
classification, and clustering \citep{Pekalska2005,Hammer2010}. Additionally, an edit distance
enables us to infer the changes we have to apply to a tree $\tre x$ to get to another tree $\tre y$,
which facilitates hints \citep{Paassen2018JEDM,Price2017,Rivers2017}. 
However, there are several subtle problems that can complicate a practical application.
First, there may exist exponentially many edit paths between $\tre x$ and $\tre y$ and it is
not always easy to choose among them, especially since many intermediate trees may be syntactically
invalid or implausible to students \citep{Paassen2018JEDM,Rivers2017}.
More generally, distances require reference programs to which distances can be computed.
This makes distance-based methods challenging to scale for larger datasets where many reference
programs exist.
By contrast, ast2vec can translate a new program into a vector without reference to past data.
Only the model parameters have to be stored, which is constant size and runtime versus linear size
and runtime. Finally, ast2vec enables us to decouple the general problem of representing
programs as vectors from task-specific problems. We can utilize the implicit information of 500,000
student programs in ast2vec in a new task and may need only little new student data to solve the
additional, task-specific problem at hand.

That being said, the basic logic of distance measures is still crucial to ast2vec. In particular,
the negative log-likelihood in Equation~\ref{eq:vae} can be interpreted as a measure of distance
between the original program and its autoencoded version, and our notion of a smooth encoding
can be interpreted as small distances between vectors in the encoding space implying small distances
of the corresponding programs.

\paragraph{Clustering:} One of the challenges of computer programs is that the space of
programs for the same task is very large and it is infeasible to design feedback for
all possible programs. Accordingly, several researchers have instead grouped programs into
a small number of clusters, for which similar feedback can be given
\citep{Choudhury2016,Glassman2015,Gross2014,Gulwani2018}. Whenever a new student requests help, one can
simply check which cluster the student's program belongs to, and assign the feedback for that
cluster. One typical way to perform clustering is by using a pairwise distance measure on
trees, such as the tree edit distance \citep{Choudhury2016,Gross2014,Zhang1989}. However, it
is also possible to use clustering strategies more specific to computer programs, such as
grouping programs by their control flow \citep{Gulwani2018} or by the unit tests they pass
\citep{McBroom2021}.
ast2vec can be seen as a preprocessing step for clustering. Once all programs are encoded as
vectors, standard clustering approaches can be applied. Additionally, ast2vec
provides the benefit of being able to interpret the clustering by translating cluster centers
back into syntax trees (refer to Section~\ref{sec:clustering}).

\paragraph{Execution Behavior:} Most representations so far focused on the
AST of a program. However, it is also possible to represent programs in terms of their execution
behavior. The most popular example is the representation by test
cases, where a program is fed with example input. If the program's output is equal to the
expected output, we say the test has passed, otherwise failed \citep{Ihantola2010}. This is
particularly useful for automatic grading of programs because test cases verify the
functionality, at least for some examples, while giving students freedom on how to
implement this functionality \citep{Ihantola2010}. Further, failing a certain unit test can
indicate a specific misconception, warranting functionality-based feedback \citep{McBroom2021}.
However, for certain types of tasks the computational path toward an output may be relevant,
for example when comparing sorting programs. \citet{Paassen2016} therefore use the
entire computation trace as a representation, i.e.\ the sequence of states of all variables in
a program.
Unfortunately, a mismatch in the output or in the execution behavior toward the output
is, in general, difficult to relate to a specific change the student would need to perform
in order to correct the problem. To alleviate this challenge, test cases have to be carefully
and densely designed, or the challenge has to be left to the student. As such, we believe that
there is still ample room for AST-based representations, like our proposed vector encodings,
which are closer to the actions a student can actually perform on their own code.

\paragraph{Neural networks:} Prior work already has investigated neural networks to encode
computer programs. In particular, \citet{Piech2015ICML} used a recursive tensor network to
encode syntax trees and performed classification on the embeddings to decide whether certain
feedback should be applied or not; \citet{Alon2019} represented syntax trees as a collection
of paths, encoded these paths as vectors and then computed a weighted average to obtain an
overall encoding of the program; \citet{Chen2018} translate programs into other programming
languages by means of an intermediary vector representation; and \citet{Dai2018} proposed an auto-encoding model for
computer programs that incorporates an attribute grammar of the programming language to not only
incorporate syntactic, but also semantic constraints (like that variables can only be used after
declaration).

Both the works of \citet{Piech2015ICML} and \citet{Alon2019} are different from our contribution
because they do not optimize an autoencoding loss but instead train an neural net for
the classification of programs, i.e.\ into feedback classes or into tags. This scheme is unable
to recover the original program from the network's output and requires expert labelling for all
training programs. Ast2vec has neither of those limitations.
The work of \citet{Chen2018} is more similar in that both input and output are programs. However,
the network does not include grammar constraints and uses an attention mechanism on the original tree to decide on its output. This is not possible
in our setting where we wish to perform datamining solely in the vector space of encodings and
be able to decode any encoding back into a tree, without reference to a prior tree.
The most similar prior work to our own is the autoencoding model of \citet{Dai2018}, which is also an autoencoding model with grammar constraints, albeit with an LSTM-based encoder and decoder. 
One could frame ast2vec as a combination of the auto-encoding ability and grammar knowledge of \citet{Dai2018} 
with the recursive processing of \citet{Piech2015ICML}, yielding a recursive tree grammar autoencoder 
\citep{Paassen2021}. That being said, future work may incorporate more recurrent network concepts and thus improve autoencoding error further.

\paragraph{Next-step hints:} Ample prior work has considered the problem of predicting the next step a student should take in a learning system, refer e.g.\ to the review of \citep{McBroom2019}. On a high level, this concerns the problem of selecting the right sequence of lessons to maximize learning gain \citep{Lan2014,Reddy2016}. In this paper, we rather consider the problem of predicting the next code change within a single programming task. This problem has been considered in more detail, for example, by \citet{Rivers2017}, \citet{Price2017}, \citet{Paassen2018JEDM}, and \citet{Price2019}. Here, we compare to two baselines evaluated by \citet{Price2019}, namely the continuous hint factory \citep{Paassen2018JEDM} and a nearest-neighbor prediction \citep{Gross2015}. We note, however, that we only evaluate the predictive accuracy, not the hint quality, which requires further analysis \citep{Price2019}.

\section{Conclusion} \label{sec:conclusion}
In this paper, we presented ast2vec, a novel autoencoder to translate the syntax trees of beginner Python 
programs to vectors in Euclidean space and back. We have trained this autoencoder on almost 500,000 student
Python programs and evaluated it in three settings. First, we utilized the network for a variety of analyses
on a classroom-sized dataset, thereby demonstrating the flexibility of the approach qualitatively.
As part of our qualitative analysis, we also showed that the encoding space of ast2vec is smooth - at least for
the example - and introduced two novel techniques for analyzing programming data based on ast2vec, namely
progress-variance-projections for a two-dimensional visualization, and a linear dynamical systems method that
guarantees convergence to the correct solution.

In terms of
quantitative analyses, we evaluated the autoencoding error of ast2vec as well as the predictive accuracy of a
simple linear model on top of ast2vec. We found that ast2vec had a low autoencoding error for the majority of 
programs, including on two unseen, large-scale datasets, though the error tended to increase with tree size.
In addition, the encoding and decoding times were low and consistent with the theoretical $\mathcal{O}(n)$ 
bounds, suggesting ast2vec is scalable. Moreover, by coupling ast2vec with our linear dynamical systems method,
we were able to approach the predictive accuracy of existing methods with a very simple model that
is both time- and space-efficient and guarantees convergence to a correct solution.

While we believe that these results are encouraging, we also acknowledge limitations: At present, ast2vec
does not decode the content of variables, although such content may be decisive to solve a task correctly.
Further improvements in terms of autoencoding error are possible as well, especially for larger programs, perhaps by including more recurrent networks as encoders and decoders.
Finally, the predictive accuracy of our proposed linear dynamical systems model has not yet achieved the
state-of-the-art and nonlinear predictors are likely necessary to improve performance further.

Still, ast2vec provides the educational datamining community with a novel tool that can be utilized without any need for further deep learning in a broad variety of analyses. We are excited
to see its applications in the future.

\section{Acknowledgements}

Funding by the German Research Foundation (DFG) under grant number PA 3460/1-1 is gratefully acknowledged.

\arxivswitch{
\bibliographystyle{acmtrans}
}{
\bibliographystyle{plainnat}
}
\bibliography{./ref}

\begin{appendix}

\section{Proofs for the dynamical system analysis}

\begin{thm}\label{thm:stability}
Let $f(\vec x) = \vec x + \bm{W} \cdot (\vec x_* - \vec x)$ for some matrix
$\bm{W} \in \R^{n \times n}$. If $|1-\lambda_j| < 1$ for all eigenvalues $\lambda_j$ of
$\bm{W}$, $f$ asymptotically converges to $\vec x_*$.

\begin{proof}
We note that the theorem follows from general results in stability analysis, especially
Lyapunov exponents \citep{Politi2013}. Still, we provide a full proof here to
give interested readers insight into how such an analysis can be done.

In particular, let $\vec x_1$ be any $n$-dimensional real vector. We know wish to show
that plugging this vector into $f$ repeatedly yields a sequences $\vec x_1, \vec x_2, \ldots$
with $\vec x_{t+1} = f(\vec x_t)$ which asymptotically converges to $\vec x_*$ in the
sense that
\begin{equation*}
\lim_{t \to \infty} \quad \vec x_t = \vec x_* .
\end{equation*}
To show this, we first define the alternative vector $\hat x_t := \vec x_t - \vec x_*$.
For this vector we obtain:
\begin{equation*}
\hat x_{t+1} = \vec x_{t+1} - \vec x_* = f(\vec x_t) - \vec x_*
= \vec x_t + \bm{W} \cdot (\vec x_* - \vec x_t) - \vec x_*
= \big(\bm{I} - \bm{W}) \cdot \hat x_t
= \big(\bm{I} - \bm{W})^t \cdot \hat x_1 .
\end{equation*}
Further, note that our desired convergence of $\vec x_t$ to $\vec x_*$ is equivalent
to stating that $\hat x_t$ converges to zero. To show that $\hat x_t$ converges to
zero, it is sufficient to prove that the matrix $\big(\bm{I} - \bm{W})^t$ converges
to zero.

To show this, we need to consider the eigenvalues of our matrix
$\bm{W}$. In particular, let $\bm{V} \cdot \bm{\Lambda} \cdot \bm{V}^{-1} = \bm{W}$
be the eigenvalue decomposition of $\bm{W}$, where $\bm{V}$ is the matrix of
eigenvectors and $\bm{\Lambda}$ is the diagonal matrix of eigenvalues
$\lambda_1, \ldots, \lambda_n$. Then it holds:
\begin{align*}
\big(\bm{I} - \bm{W})^t
&= \big(\bm{V} \cdot \bm{V}^{-1} - \bm{V} \cdot \bm{\Lambda} \cdot \bm{V}^{-1}\big)^t
= \big(\bm{V} \cdot (\bm{I} - \bm{\Lambda}) \cdot \bm{V}^{-1}\big)^t \\
&= \bm{V} \cdot (\bm{I} - \bm{\Lambda}) \cdot \bm{V}^{-1} \cdot \bm{V} \cdot \ldots \cdot
\bm{V}^{-1} \cdot \bm{V} \cdot (\bm{I} - \bm{\Lambda}) \cdot \bm{V}^{-1}
= \bm{V} \cdot (\bm{I} - \bm{\Lambda})^t \cdot \bm{V}^{-1} .
\end{align*}
Now, let us consider the matrix power $(\bm{I} - \bm{\Lambda})^t$ in more detail.
Because this is a diagonal matrix, the power is just applied elementwise.
In particular, for the $j$th diagonal element we obtain the power $(1 - \lambda_j)^t$.
In general, the $j$th eigenvalue can be complex-valued. However, the absolute value
still behaves like $|(1-\lambda_j)^t| = |1-\lambda_j|^t$. Now, since we required
that $|1-\lambda_j| < 1$ we obtain
\begin{equation*}
\lim_{t \to \infty} \quad |1-\lambda_j|^t = 0 \qquad \Rightarrow \qquad
\lim_{t \to \infty} \quad \bm{V} \cdot (\bm{I} - \bm{\Lambda})^t \cdot \bm{V}^{-1} = \bm{0},
\end{equation*}
which implies our desired result.
\end{proof}
\end{thm}

\begin{thm}\label{thm:linreg}
Problem~\ref{eq:dynsys_learning} has the unique solution in Equation~\ref{eq:linreg}.

\begin{proof}
We solve the problem by setting the gradient to zero and considering the Hessian.

First, we compute the gradient of the loss with respect to $\bm{W}$.
\begin{align*}
&\nabla_{\bm{W}} \sum_{t=1}^{T-1} \lVert \vec x_t + \bm{W} \cdot (\vec x_* - \vec x_t) - \vec x_{t+1}\rVert^2 + \lambda \cdot \lVert \bm{W} \rVert^2_\mathcal{F} \\
=& 2 \cdot \sum_{t=1}^{T-1} \big(\vec x_t + \bm{W} \cdot (\vec x_* - \vec x_t) - \vec x_{t+1}\big) \cdot (\vec x_* - \vec x_t)^T
+ 2 \cdot \lambda \cdot \bm{W}
\end{align*}
Setting this to zero yields
\begin{align*}
\bm{W} \cdot \Big(\sum_{t=1}^{T-1} (\vec x_* - \vec x_t) \cdot (\vec x_* - \vec x_t)^T + \lambda \cdot \bm{I}\Big)
&= \sum_{t=1}^{T-1} (\vec x_t - \vec x_{t+1}) \cdot (\vec x_* - \vec x_t)^T \\
\iff \quad \bm{W} \cdot \big( \bm{X}_t^T \cdot \bm{X}_t + \lambda \cdot \bm{I}\big)
&= \big(\bm{X}_{t+1}-\bm{X}_t \big)^T \cdot \bm{X}_t
\end{align*}
For $\lambda > 0$, the matrix $\bm{X}_t^T \cdot \bm{X}_t + \lambda \cdot \bm{I}$ is quaranteed
to be invertible, which yields our desired solution.

It remains to show that the Hessian for this problem is positive definite. Re-inspecting the
gradient, we observe that the matrix $\bm{W}$ occurs only as a product with the term
$\sum_{t=1}^{T-1} (\vec x_* - \vec x_t) \cdot (\vec x_* - \vec x_t)^T$, which is positive
semi-definite, and the term $\lambda \cdot \bm{I}$, which is strictly positive definite.
Hence, our problem is convex and our solution is the unique global optimum.
\end{proof}
\end{thm}

\end{appendix}

\end{document}